\documentclass[twoside,11pt]{article}

\usepackage{subcaption}

\usepackage[nohyperref]{jmlr2e}
\usepackage{url}
\usepackage{amsmath, amssymb}
\usepackage{xcolor}
\usepackage{booktabs}
\usepackage{ulem}
\usepackage{xspace}

\usepackage[utf8]{inputenc}
\usepackage{tikz, pgfplots, pgfplotstable}
\usepgfplotslibrary{groupplots}
\pgfplotsset{grid style=none}
\pgfplotsset{axis y line=left}
\pgfplotsset{axis x line=bottom}
\usetikzlibrary{external}
\usetikzlibrary{pgfplots.colormaps}
\definecolor{blue538}{RGB}{0, 143, 213}
\definecolor{red538}{RGB}{252, 79, 48}
\definecolor{yellow538}{RGB}{229, 174, 56}
\definecolor{green538}{RGB}{219, 219, 219}

\definecolor{v1}{rgb}{0.7607843137254902, 0.9764705882352941, 0.32941176470588235}
\definecolor{v2}{rgb}{0.6274509803921569, 0.8, 0.27058823529411763}
\definecolor{v3}{rgb}{0.4745098039215686, 0.7490196078431373, 0.43137254901960786}
\definecolor{v4}{rgb}{0.33725490196078434, 0.7411764705882353, 0.6}
\definecolor{v5}{rgb}{0.27450980392156865, 0.7607843137254902, 0.7647058823529411}
\definecolor{v6}{rgb}{0.21568627450980393, 0.7607843137254902, 0.8784313725490196}
\definecolor{v7}{rgb}{0.30980392156862746, 0.6196078431372549, 0.8156862745098039}
\definecolor{v8}{rgb}{0.38823529411764707, 0.43529411764705883, 0.6941176470588235}
\definecolor{v9}{rgb}{0.3764705882352941, 0.30196078431372547, 0.596078431372549}
\definecolor{v10}{rgb}{0.2549019607843137, 0.21568627450980393, 0.5058823529411764}

\pgfplotscreateplotcyclelist{viridis}{
{thick, v10},
{thick, v9},
{thick, v8},
{thick, v7},
{thick, v6},
{thick, v5},
{thick, v4},
{thick, v3},
{thick, v2},
{thick, v1},
}

\newcommand{\cov}{\textrm{cov}}
\newcommand{\var}{\textrm{var}}
\newcommand{\bJ}{\mathbf{J}}
\newcommand{\Kuu}{\mathbf{K}_{\mathbf{uu}}}
\newcommand{\Kuf}{\mathbf{K}_{\mathbf {uf}}}
\newcommand{\Kfu}{\mathbf{K}_{\mathbf {fu}}}
\newcommand{\Kff}{\mathbf{K}_{\mathbf {ff}}}
\newcommand{\Kpp}{\mathbf{K}_{\mathbf {\phi \phi}}}
\newcommand{\ba}{\mathbf{a}}
\newcommand{\bb}{\mathbf{b}}
\newcommand{\bw}{\mathbf{w}}
\newcommand{\bK}{\mathbf{K}}
\newcommand{\bL}{\mathbf{L}}
\newcommand{\bI}{\mathbf{I}}
\newcommand{\bff}{\mathbf f}
\newcommand{\bk}{\mathbf k}
\newcommand{\bm}{\mathbf m}
\newcommand{\bX}{\mathbf X}
\newcommand{\by}{\mathbf y}
\newcommand{\bx}{\mathbf x}
\newcommand{\bu}{\mathbf u}
\newcommand{\bv}{\mathbf v}
\newcommand{\bR}{\mathbf R}
\newcommand{\bs}{\mathbf s}
\newcommand{\bS}{\mathbf S}
\newcommand{\bZ}{\mathbf Z}
\newcommand{\bzero}{\mathbf 0}
\newcommand{\dee}{\,\textrm d}
\newcommand{\given}{\,|\,}
\newcommand{\cF}{\mathcal{F}}
\newcommand{\cH}{\mathcal{H}}
\newcommand{\cGP}{\mathcal{GP}}
\newcommand{\cP}{\mathcal{P}}
\newcommand{\cN}{\mathcal{N}}
\newcommand{\cO}{\mathcal{O}}

\newcommand\PS[1]{{\left \langle #1 \right \rangle}_{\! \cH}}
\newcommand\N[1]{{\left|\left| #1 \right|\right|}_{\! \cH}}
\newcommand\PSi[2]{{ \left \langle #1 \right \rangle}_{\! #2}}

\newcommand{\bphi}{\boldsymbol\phi}
\newcommand{\balpha}{\boldsymbol\alpha}
\newcommand{\bbeta}{\boldsymbol\beta}
\newcommand{\bSigma}{\boldsymbol\Sigma}
\newcommand{\infint}{\int_{-\infty}^\infty}
\newcommand{\kron}{\raisebox{1pt}{\ensuremath{\:\otimes\:}}}

\newcommand{\matern}{{Mat\'ern}\xspace}
\newcommand{\maternonetwo}{{Mat\'ern-\ensuremath{\tfrac{1}{2}}}\xspace}
\newcommand{\maternthreetwo}{{Mat\'ern-\ensuremath{\tfrac{3}{2}}}\xspace}
\newcommand{\maternfivetwo}{{Mat\'ern-\ensuremath{\tfrac{5}{2}}}\xspace}

\newcommand{\todo}[1]{{\color{red} TODO: #1}}
\newcommand{\james}[1]{{\color{blue} James:``#1''}}
\newcommand{\arno}[1]{{\color{blue} Arno:``#1''}}
\newcommand{\nicolas}[1]{{\color{blue} Nicolas:``#1''}}
\newlength\figurewidth
\newlength\figureheight

\newcommand\add[1]{#1}
\newcommand\remove[1]{}

\jmlrheading{ }{ }{ }{ }{ }{James Hensman, Nicolas Durrande, and  Arno Solin}

\ShortHeadings{Variational Fourier Features}{Hensman, Durrande, and Solin}
\firstpageno{1}

\makeatletter
\def\@starteditor{\noindent \small {}}
\makeatother

\begin{document}

\title{Variational Fourier Features for Gaussian Processes}

\author{\name James Hensman\thanks{This work was undertaken whilst JH was affiliated to the Centre for Health Information, Computation, and Statistcs (CHICAS), Faculty of Health and Medicine, Lancaster University}\email james@prowler.io\\
       \addr PROWLER.io\\
       66-68 Hills Road\\
       Cambridge, CB2 1LA, UK
       \AND
       \name Nicolas Durrande \email nicolas.durrande@emse.fr\\
       \addr Ecole des Mines\\
       42000 St Etienne\\
       France
       \AND
       \name Arno Solin \email arno.solin@aalto.fi\\
       \addr Aalto University \\
       FI-00076 AALTO \\
       Espoo, Finland}

\editor{}

\maketitle
\thispagestyle{plain} 
\begin{abstract}    This work brings together two powerful concepts in Gaussian processes: the
    variational approach to sparse approximation and the spectral
    representation of Gaussian processes. This gives rise to an approximation
    that inherits the benefits of the variational approach but with the
    representational power and computational scalability of spectral
    representations. The work hinges on a key result that there exist spectral
    features related to a finite domain of the Gaussian process which exhibit
    almost-independent covariances. We derive these expressions for \matern
    kernels in one dimension, and generalize to more dimensions using kernels
    with specific structures. Under the assumption of additive Gaussian noise,
    our method requires only a single pass through the dataset, \add{making for} very
    fast and accurate computation.  We fit a model to 4 million training points
    in just a few minutes on a standard laptop. With non-conjugate likelihoods,
    our MCMC scheme reduces the cost of computation from $\cO(NM^2)$ (for a
    sparse Gaussian process) to $\cO(NM)$ per iteration, where $N$ is the
    number of data and $M$ is the number of features.
\end{abstract}

\begin{keywords}
  Gaussian processes, Fourier features, variational inference
\end{keywords}
\section{Introduction}
Efficient computation in Gaussian process (GP) models is of broad interest across machine learning and statistics, with applications in \add{the fields of} spatial epidemiology \citep{diggle2013statistical, banerjee2014hierarchical}, robotics and control \citep{deisenroth2015gaussian}, signal processing \citep[e.g.][]{Sarkka2013spatiotemporal}, Bayesian optimization and probabilistic numerics, \citep[e.g.][]{osborne2010bayesian, briol2016probabilistic, hennig2015probabilistic}, and many others.

Computation is challenging for several reasons. First, the computational complexity usually scales cubically with the number of data $N$, as one must decompose a $N \times N$ dense covariance matrix. Second, the posterior is both high-dimensional and non-Gaussian\remove{, where the data likelihood, conditioned on the process, is not conjugate} \add{if the data likelihood is not conjugate}. Third, GP models have a hierarchical structure, with the latent values of the function being conditioned on parameters of the covariance function.

In this work, we adopt a variational Bayesian framework for dealing with these issues. \add{By this we mean that we minimize the Kullback--Leibler (KL) divergence $\textsc{KL}[q||p]$, where $q$ denotes an approximation to the posterior, and $p$ is the posterior itself. Many authors have applied this framework to GP problems before: perhaps the earliest example is \citet{csato2002tap}. \citet{seeger2003bayesian} also made use of this KL criterion in the context of Gaussian processes. Particularly influentuial works include \citet{titsias2009variational}, who developed the first sparse GP using the KL divergence, and \citet{opper2009variational}, who made a case for this approximation in the non-conjugate (but not sparse) case.}

The variational framework for Gaussian processes has several advantages. First, the three challenges mentioned above can be tackled by a single unified objective. Second, the accuracy of the approximation can easily be shown to increase monotonically with increasing complexity of the proposed approximating family (i.e.\ more inducing points is always better). Third, the framework provides an evidence lower bound (or ELBO), which can be evaluated to compare different approximations.

In this work, we combine the variational approach with Fourier features; we refer to our method as Variational Fourier Features (VFF).  Whilst most sparse Gaussian process approximations rely on {\it inducing} or {\it pseudo} inputs, which lie in the same domain as the inputs to the GP, Fourier features lie in the spectral domain of the process. As inducing-point approximations build up the posterior through kernel functions, Fourier features build up the approximate posterior through sinusoids. Our approach differs from existing works which use random frequencies \citep{rahimi2007random, rahimi2009weighted} or optimized frequencies \citep{lazaro2010sparse}, in that we use frequencies placed on a regular grid \citep[cf.][]{solin2014hilbert}, which is key in our derivation of a computationally convenient matrix structure.

Naive combination of Fourier features with sparse GP approaches is not feasible because the standard Fourier transform of a stochastic process does not converge, meaning that these features are associated with random variables of infinite variance. \citet{matthews2015sparse} have shown that valid inducing features for the variational approach require finite, deterministic projections of the process. To combat this, \citet{figueiras2009inter} used a windowing technique, which made the integrals converge to valid inducing features (for the squared-exponential kernel), but lost perhaps the most important aspect of a Fourier approach: that the features should be independent. In this work, we combine a finite window with a Reproducing Kernel Hilbert Space (RKHS) projection to construct features that are {\it almost} independent, having covariance structures that are easily decomposed. This gives rise to a complexity of $\mathcal O(NM)$, comparing favourably with the $\mathcal O(NM^2)$ complexity of other variational approaches, where $M$ denotes the number of features.

The contributions of this paper are the following:
\begin{itemize}
  \item We introduce a novel approximation scheme for representing a GP using a Fourier decomposition. We limit our interest to the widely used \matern family of stationary kernels, for which our proposed decomposition exhibits an almost-independent structure.
  \item We combine our approximation scheme with the variational Bayesian framework for Gaussian processes, and present how the methodology generalizes to GP inference problems with general likelihoods.
\end{itemize}
Consequently, this paper aims to provide an approximate Gaussian process inference scheme which is theoretically sound, has strong representative power, is extremely fast, and---most importantly---works well in practice.

The rest of this paper is structured as follows. Section~\ref{par:background} reviews the relevant background related to Gaussian process inference and previous work in sparse and spectral approximations to GP models. In Section~\ref{par:vff} the core methodology for Variational Fourier Features is presented for the one-dimensional case. In Section \ref{par:extending} this is extended to multidimensional cases with additive and product structures. Implementation details and computational complexity of the method are covered in Section~\ref{par:details}. Section~\ref{par:experiments} is dedicated to a set of illustrative toy examples and a number of empirical experiments, where practical aspects of Variational Fourier Feature inference are demonstrated. Finally, we conclude the paper with a discussion in Section~\ref{par:discussion}.

 \section{Background}\label{par:background}
In this section, we first set up the family of Gaussian process models that we will consider, which allows us to establish some notation. Next, we review some basic results regarding the spectrum of stationary Gaussian processes, and recall how Fourier features approximate the kernel. \add{We relate sparse Gaussian process approximations and Fourier approximations by explicating them as alternative models \citep{quinonero2005unifying}. We then recap the variational approximation to Gaussian processes, including expressions for sparse approximations and approximations for non-conjugate likelihoods. }The two final subsections in this section discuss decomposition of Gaussian processes: we first link decomposition and conditioning and then discuss inter-domain decomposition.

\subsection{Gaussian process models}
Gaussian processes are distributions over functions, defined by a mean function and a covariance function \citep[see][for an introduction]{williams2006gaussian}. In this section we consider GPs over the real line, $x\in \mathbb R$. Making the standard assumption that the mean function is zero, we write
\begin{equation}
    f(x) \sim \mathcal{GP}\left(0,\,k(x, x')\right)\,.
\end{equation}
The data $\by = [y_n]_{n=1}^N$ at locations $\bX = [x_n]_{n=1}^N$ are conditioned on the function evaluations $\bff = [f(x_n)]_{n=1}^N$ through some factorising likelihood
\begin{equation}
    p(\by \given f(x)) = p(\by\given\bff) = \prod_n p(y_n\given f(x_n))\,,
\end{equation}
which we do not in general assume to be Gaussian.
A defining property of Gaussian processes is that any finite number of function evaluations follow a multivariate normal distribution, so
\begin{equation}
    p(\bff) = \mathcal N(\mathbf{0}, \, \Kff)\,,
\end{equation}
and the process conditioned on these values is given by a conditional Gaussian process\remove{ which we call $f^\star(x)$}:
\begin{equation}
f(x) \given \bff \sim \mathcal {GP}\left(\bk_\bff(x)\add{^\top} \Kff^{-1}\bff, \; k(x, x') - \bk_\bff(x)^\top \Kff^{-1}\bk_\bff(x')\right)\,.
\label{eq:gp_cond}
\end{equation}

To compute the posterior over functions given the data $f(x)\given \by$, it is possible to compute the posterior only for the finite set $p(\bff\given\by)$ using standard methodologies like Markov Chain Monte Carlo (MCMC) \citep[e.g.][]{murray2010elliptical,filippone2013comparative} or variational Bayes \citep[e.g.][]{opper2009variational, khan2012fast, khan2013fast}. Evaluating the function at a test point means averaging equation~\eqref{eq:gp_cond} over this posterior $p(\bff\given \by)$.

In our notation, the matrix $\Kff$ is given by evaluating the covariance function at all pairs of data points $\Kff{[i, j]} = k(x_i, x_j)$. The vector valued function $\bk_\bff(x)$ is given similarly, $\bk_\bff(x) = [k(x_1, x), k(x_2, x), \ldots, k(x_n, x)]^\top$. We omit dependence on covariance function parameters\remove{ and inputs $\bX$} for clarity.

\subsection{Spectral representations of stationary covariances}
Stationary Gaussian processes are those with a covariance function that can be written as a function of the distance between observations, $k(x, x') = k(|x-x'|) = k(r)$. Of particular interest in this work will be the \matern family with half-integer orders, the first three of which are
\begin{equation}
\begin{aligned}
    k_{\tfrac{1}{2}}(r) &= \sigma^2 \exp( -r / \ell )\,,\\
    k_{\tfrac{3}{2}}(r) &= \sigma^2 (1 + \sqrt{3} r / \ell)\exp( - \sqrt{3} r / \ell) \,,\\
    k_{\tfrac{5}{2}}(r) &= \sigma^2 (1 + \sqrt{5} r / \ell + \tfrac{5}{3} r^2/\ell^2)\exp( - \sqrt{5} r / \ell)\,,
\end{aligned}
\end{equation}
where $\sigma^2$ and $\ell$ are variance and lengthscale parameters respectively.

Bochner's theorem \citep{Akhiezer+Glazman:1993} tells us that any continuous positive definite function, such as a covariance function, can be represented as \remove{a}\add{ the} Fourier transform of a positive measure. If the measure has a density, it is known as the spectral density $s(\omega)$ of the covariance function. This gives rise to the Fourier duality of spectral densities and covariance functions, known as the Wiener-Khintchin theorem. It gives the following relations:
\begin{alignat}{3}
    & s(\omega) &&= \mathcal F\{k(r)\} &&= \int_{-\infty}^\infty k(r) e^{-i\omega r} \dee r\,\add{,}\\
    & k(r) &&= \mathcal F^{-1}\{s(\omega)\} &&= \frac{1}{2\pi} \int_{-\infty}^\infty s(\omega) e^{i\omega r} \dee \omega\,\label{eq:inv_fourier}.
\end{alignat}
Since kernels are symmetric\remove{,} real functions, the spectrum of the process is also a symmetric\remove{,} real function. The spectra corresponding to the \matern covariances are
\begin{alignat}{3}
    &s_{\tfrac{1}{2}}(\omega) &&= 2\sigma^2\lambda (\lambda^2 + \omega^2)^{-1},&& \qquad\lambda=\ell^{-1},\\
    &s_{\tfrac{3}{2}}(\omega) &&= 4\sigma^2\lambda^3 (\lambda^2 + \omega^2)^{-2},&& \qquad\lambda=\sqrt{3}\ell^{-1},\\
    &s_{\tfrac{5}{2}}(\omega) &&= \tfrac{16}{3}\sigma^2\lambda^5 (\lambda^2 + \omega^2)^{-3},&& \qquad\lambda=\sqrt{5}\ell^{-1}\,.
\end{alignat}

\subsection{Model approximations}
In order to overcome the cubic computational scaling of Gaussian process methods, many approximations have been proposed. Here we \add{briefly} review two types of approximations, those based on the spectrum of the process and \remove{so-called} \add{others referred to as } `sparse' approximations.
\subsubsection{Random Fourier features}
\label{par:rff}
Random Fourier Features (RFF) is a method for approximating kernels.
The essential element of the RFF approach \citep{rahimi2007random, rahimi2009weighted} is the realization that the Wiener-Khintchin integral \eqref{eq:inv_fourier} can be approximated by a Monte Carlo sum
\begin{equation}
    k(r) \approx\tilde k(r) = \frac{\sigma^2}{M}\sum_{m=1}^M \cos(\omega_m r),
\end{equation}
where the frequencies $\omega_m$ are drawn randomly from the distribution \add{proportional to}  $s(\omega)$.
Note that the imaginary part  is zero because $s(\omega r)$ is even and $i \sin(\omega r)$ is odd.

This approximate kernel has a finite basis function expansion as
\begin{equation}
    \bphi(x) = [\cos(\omega_1 x), \cos(\omega_2 x), \ldots, \cos(\omega_M x), \sin(\omega_1 x), \sin(\omega_2 x), \ldots, \sin(\omega_M x)]^\top\, ,
    \label{eq:basis_expansion}
\end{equation}
and we can write the approximate Gaussian process as
\begin{align}
    f(x) \sim \mathcal {GP} \left(0,\, \tilde k(x-x')\right)\qquad \text{or}\qquad f(x) \sim \mathcal{GP}\left(0, \, \frac{\sigma^{2}}{M}\bphi(x)^\top\bphi(x')\right) 
\end{align}
or equivalently as a parametric model, in the form
\begin{align}
    f(x)  &= \bphi(x)^\top\bw, \\
    \bw &\sim \mathcal N \left( \bzero, \frac{\sigma^2}{M}\bI \right)\,.
    \label{eq:rff_model}
\end{align}

\subsubsection{Optimized Fourier features}
\label{par:optff}
In order to have a representative sample of the spectrum, the RFF methodology typically requires the number of spectral sample points to be large. In \citet{lazaro2010sparse}, this was addressed by the attractive choice of optimizing the spectral locations along with the hyperparameters (the `Sparse Spectrum GP'). However, as argued by those authors, this option does not converge to the full GP and can suffer from overfitting to the training data. We confirm this empirically in a simple experiment in Section \ref{par:experiments}.

\add{Noting that this method was prone to overfitting, \citet{gal2015improving} sought to improve the model by integrating out, rather than optimizing the frequencies. Gal and Turner derived a variational approximation that made use of a tractable integral over the frequency space. The result is an algorithm that suffers less overfitting than the Sparse Spectrum GP, yet remains flexible. We emphasize the difference to the approach in this work: Gal and Turner  proposed variational inference in a sparse spectrum model that is derived from a GP model; our work aims to directly approximate the posterior of the true models using a variational representation. }

\subsubsection{Regular Fourier features}
Another way to approximate the integral in \eqref{eq:inv_fourier} is with a regularly spaced set of frequencies $\omega$. We refer to this method as {\it regular} Fourier features. While the Random Fourier Features approach could be interpreted as a Monte Carlo approximation to the Wiener-Khintchin integral, the regular Fourier features approach can be seen as a quadrature approximation of the same integral. Therefore the regular methods are not {\it sparse} in spectral sense, but {\it dense} and deterministic given the domain of interest \citep{solin2014hilbert}.

The approximation to the original covariance function takes the form of a finite sum
\begin{equation}
    k(r) \approx\tilde k(r) = \sum_m s(\omega_m)\cos(\omega_m r)\qquad \omega_m = m \Delta,
\end{equation}
where $\Delta$ is a small number which defines the grid spacing. Similarly to Random Fourier features, the resulting approximate Gaussian process can be written as a parametric model:
\begin{align}
    f(x)  &= \bphi(x)^\top\bw, \\
    \bw &\sim \mathcal N(\bzero,\, \mathbf{S}) \qquad \mathbf{S} = \text{diag}(s(\omega_1),\ldots,s(\omega_M),s(\omega_1),\ldots,s(\omega_M))\,.
\end{align}
Like in the Random Fourier Features approximation \eqref{eq:rff_model}, regular Fourier features are independent (the matrix $\mathbf S$ is diagonal),
  and if the inputs are spaced on a regular grid, fast Fourier transform (FFT) methods can speed up the computations to $\mathcal O(N \log N)$ \citep[see e.g.][]{Paciorek:2007, Fritz+Neuweiler+Nowak:2009}.

\subsubsection{Sparse Gaussian process models}
In the above we reviewed how spectral approximations result in parametric models. In their review paper, \citet{quinonero2005unifying} also presented a view of sparse Gaussian processes where the result is an approximate, parametric model.

The terminology `sparse' originally referred to methods which restricted computation to a subset of the data points \citep[e.g.][]{csato2002sparse, williams2001using}, until \citet{snelson2005sparse} relaxed this assumption with the idea of {\em pseudo inputs}.
The pseudo-inputs (or `inducing points') $\bZ=[z_m]_{m=1}^M$ lie in the same domain as $\bX$, but with $M < N$. The values of the process associated with these points are denoted $\bu = [f(z_m)]_{m=1}^M$.
The simplest form of sparse Gaussian process model using these variables is the Deterministic Training Conditional (DTC) model, written
\begin{align}
    f(x)  &= \bk_\bu(x)^\top\Kuu^{-1}\bu, \\
    \bu &\sim \mathcal N(\bzero, \Kuu)\,.
    \label{eq:dtc_model}
\end{align}
This kind of `projected process' approximation has also been discussed by e.g. \citet[][]{banerjee2008gaussian}. The literature contains many parametric models that approximate Gaussian process behaviours; for example \citet{bui2014} included tree-structures in the approximation for extra scalability, and \citet{moore2015} combined local Gaussian processes with Gaussian random fields.

\subsection{Variational Gaussian process approximations}
\label{par:sparseGP}
We have seen how Fourier approximations and sparse Gaussian processes can be written as approximate parametric models.
The variational approximation to Gaussian processes provides a more elegant, flexible and extensible solution in that the {\it posterior distribution} of the original model is approximated, rather than the model itself. In existing work, the variational approach has been used alongside ideas from the sparse GP literature: the remit of this work is to combine the variational methodology with Fourier based approximations.  We provide a short review of the variational methodology here, for a more detailed discussion see \citet{matthews2016scalable}.

In variational Bayes \citep[see e.g.][for a contemporary review]{blei2016variational} the idea is to approximate the posterior of a model by selecting the optimal distribution from a fixed family. Optimality is usually defined through the Kullback-Leibler divergence
\begin{equation}
    \textsc{KL}[q(f(x)) \,\|\, p(f(x)\given \by)] = \mathbb E_{q(f(x))} \left[\log q(f(x)) - \log p(f(x)\given \by)\right]\,.
    \label{eq:KL_def}
\end{equation}
This equation is a slight abuse of notation, since it is not legitimate to write $p(f(x))$. Nonetheless, the intuition is good, and our overview has the same result as given by a more technical derivation \citep{matthews2015sparse}.

\subsubsection{The approximating stochastic process}
We are tasked with defining a family of variational distributions $q$. Similarly to the sparse GP models, we introduce a set of pseudo inputs (or `inducing points') $\bZ=[z_m]_{m=1}^M$, which lie in the same domain as $\bX$, but with $M < N$.  Note that $\bZ$ do not form part of the model, but only the variational approximating distribution $q(f(x))$: we may choose them freely, and we assume here that $\bX$ and $\bZ$ do not intersect. We collect the values of the function at $\bZ$ into a vector $\bu = [f(z_m)]_{m=1}^M$, and write the process conditioned on these values as
\begin{equation}
    f(x) \given \bu \sim \mathcal{GP}\left(\bk_\bu(x)\add{^\top} \Kuu^{-1}\bu, \; k(x, x') - \bk_\bu(x)^\top \Kuu^{-1}\bk_\bu(x')\right)\,,
\label{eq:gp_cond_sparse}
\end{equation}
echoing the true posterior of the process, equation~\eqref{eq:gp_cond}. We emphasize that we have {\it selected} this particular form of distribution for $q(f(x)\given \bu)$, and the joint approximation for the posterior is $q(\bu)q(f(x)\given\bu)$. Whilst most of the flexibility of the approximation comes from $q(\bu)$, using this form of conditional GP for $q(f(x) \given\bu)$ means that the approximation cannot be written as a parametric model, and does not suffer the degenerate behaviour associated with the DTC approximation \eqref{eq:dtc_model}.

We must also choose some approximate posterior distribution $q(\bu)$, whose exact form will depend on the likelihood $p(\by\given f(x))$ as we shall discuss momentarily.

\subsubsection{The \textsc{ELBO}}
Variational Bayes proceeds by maximizing the Evidence Lower Bound (ELBO), which indirectly  {\it minimizes} the Kullback-Leibler objective. By expanding the true posterior using Bayes' rule, and substituting this into equation \eqref{eq:KL_def}, we have
\begin{align}
    \textsc{KL}\left[q(f(x))\,\|\,p(f(x)\given\by)\right] &= -\mathbb E_{q(f(x))}\left[ \log \frac{p(\by\given f(x)) \, p(f(x))}{q(f(x))}\right] + \log p(\by)\nonumber\\
                                                          &\triangleq -\textsc{ELBO} + \log p(\by)\,.
    \label{eq:elbo}
\end{align}
To obtain a tractable \textsc{ELBO} for Gaussian processes, we factor both prior and approximate posterior processes into conditional GPs, conditioning on: the inducing input points and values $(\bZ, \bu)$; the data pairs $(\bX, \bff)$; the remainder of the process $f(x)$. The prior distribution on the process $p(f(x))$ can be written $p(\bu)p(\bff\given \bu)p(f\given \bff,\bu)$, with
\begin{equation}
\begin{aligned}
    p(\bu) &= \mathcal N\left(\mathbf{0},\, \Kuu\right)\\
    p(\bff\given\bu) &= \mathcal N\left(\Kfu\Kuu^{-1}\bu,\, \Kff - \Kfu\Kuu^{-1}\Kfu^\top\right)\\
    p(f(x)\given \bff,\bu) &= \mathcal {GP}\left(m^\star(x),\,k^\star(x, x')\right)\,.
\end{aligned}
\end{equation}
where $m^\star(x)$ and $k^\star(x, x')$ are the usual Gaussian process conditional mean and variance, conditioned on both $\bff$ and $\bu$.
The approximate posterior process can be factored similarly:
\begin{equation}
\begin{aligned}
        q(\bff\given\bu) &= \mathcal N\left(\Kfu\Kuu^{-1}\bu,\, \Kff - \Kfu\Kuu^{-1}\Kfu^\top\right)\, ,\\
    q(f(x)\given \bff,\bu) &= \mathcal {GP}\left(m^\star(x),\,k^\star(x, x')\right)\,.
\end{aligned}
\end{equation}
Noting that the two processes are the same, aside for $p(\bu)$ and $q(\bu)$, substituting into equation \eqref{eq:elbo} significantly simplifies the \textsc{ELBO}:
\begin{equation}
    \textsc{ELBO} = \mathbb E_{q(\bu)q(\bff\given\bu)} \big[\log p(\by\given\bff)\big] - \mathbb E_{q(\bu)}\left[\log\frac{q(\bu)}{p(\bu)}\right]\,.
    \label{eq:gp_elbo}
\end{equation}

\subsubsection{The optimal approximation} It is possible to show that the distribution $\hat q(\bu)$ that maximizes the \textsc{ELBO} is given by
\begin{equation}
    \log \hat q(\bu) = \mathbb E_{q(\bff\given \bu)} \left[\log p(\by\given\bff)\right] + \log p(\bu) + \textit{const.}
    \label{eq:q_hat}
\end{equation}
This distribution is intractable for general likelihoods, but can be well approximated using MCMC methods.
Parameters of the covariance function can be incorporated into the MCMC scheme in a straightforward manner, see \citet{hensman2015mcmc} for details. 

\subsubsection{Gaussian approximations}
\label{par:gaussian_approximations}
A computationally cheaper method than sampling $\hat q(\bu)$ is to approximate it using a Gaussian distribution, and to optimize the \textsc{ELBO} with respect to the mean and covariance of the approximating distribution, in order to minimize the KL divergence \citep{hensman2015scalable}. If the approximating Gaussian distribution $q(\bu)$ has mean $\bm$ and covariance $\bSigma$, then the entire approximating process is a GP, with
\begin{align}
    q(f(x)) &= \int q(\bu)q(f(x)\given \bu)\dee \bu \nonumber \\
            &= \mathcal {GP}\left(\bk_\bu(x)\add{^\top} \Kuu^{-1}\bm, \, k(x, x') + \bk_\bu(x)^\top (\Kuu^{-1}\bSigma\Kuu^{-1} - \Kuu^{-1})\bk_\bu(x')\right)\,.
\label{eq:gp_pred_sparse}
\end{align}

The covariance function parameters can be estimated by optimization of the \textsc{ELBO} alongside the variational parameters, which leads to approximate maximum likelihood estimation. This may lead to some bias in the estimation of the parameters \citep{turner2011two}, but the method is reported to work well in practice \citep{chai2012variational, hensman2015scalable, lloyd2015variational, dezfouli2015scalable}.

\subsubsection{Gaussian likelihood}
For the special case where the data-likelihood $p(y_n\given f(x_n))$ is Gaussian with noise-variance $\sigma_n^2$, the optimal distribution for $q$ is given by
\begin{equation}
\begin{aligned}
    \hat q(\bu) = \mathcal N\left(\hat\bm,\, \hat\bSigma \right),
    \quad\text{with}\quad
  & \hat\bSigma = [\Kuu^{-1} + \sigma_n^{-2}\Kuu^{-1}\Kuf\Kuf^\top\Kuu^{-1}]^{-1}\,, \\
  & \hat\bm=\sigma_n^{-2}\hat\bSigma\Kuf\by\,,
\end{aligned}
\label{eq:optimal_q}
\end{equation}
and the \textsc{ELBO} at this optimal point is
\begin{equation}
    \textsc{ELBO}\left(\hat q\right) = \log \mathcal N\left(\by \given 0,\, \Kfu\Kuu^{-1}\Kuf + \sigma_n^2 \mathbf {I}\right) - \tfrac{1}{2}\sigma_n^{-2} \textrm{tr}\!\left(\Kff - \Kfu\Kuu^{-1}\Kuf\right)\,.
\end{equation}
This expression might give the misleading impression that this approximation resembles the DTC method \citep[described by][]{quinonero2005unifying} if one interprets the matrix $\Kfu\Kuu^{-1}\Kuf$ as an approximate prior. However, prediction in the variational approximation does not suffer the same degenerate behaviour as DTC \citep{titsias2009variational, hensman2013gaussian}.

\subsubsection{Prediction}
Prediction in the variational approximation to Gaussian processes differs from the view given in \citet{quinonero2005unifying}, and it is here that the elegance of the method is apparent. It is {\em not} necessary to make any additional approximations at predict-time,  since the whole process has already been approximated. For Gaussian posteriors (and Gaussian approximate posteriors), one simply evaluates the GP (equation~\eqref{eq:gp_pred_sparse}). For non Gaussian approximations to $q(\bu)$, one must average the conditional equation~\eqref{eq:gp_cond_sparse} under $q(\bu)$.

\subsection{Linking conditioning and decomposition}
\label{par:projection}
The sparse-variational methodology that we have presented involved factoring the Gaussian process as $p(f(x)) = p(f(x)\given \bu)p(\bu)$. A related and useful concept is decomposition of the process: we will make use of such a decomposition in sebsequent sections to derive Variational Fourier Features.

Let $f(x)\sim \mathcal {GP}\big(0, k(x, x')\big)$. We can decompose $f$ as the sum of two independent Gaussian processes:
\begin{align}
    g(x) &\sim \mathcal{GP}\big(0,\, \bk_\bu(x)^\top\bK_{\bu\bu}^{-1}\bk_\bu(x')\big)\,,\\
    h(x) &\sim \mathcal{GP}\big(0,\, k(x, x') - \bk_\bu(x)^\top\bK_{\bu\bu}^{-1}\bk_\bu(x')\big)\,,\\
    f(x) &= g(x) + h(x).
    \label{eq:gp_project1}
\end{align}
It is clear from the additivity of the covariance functions that this sum recovers the correct covariance for $f(x)$. Note that the covariance of $h(x)$ is the same as for $p(f(x)\given \bu)$. We say that $g(x)$ is the projection of $f(x)$ onto $\bu$, and $h(x)$ is the orthogonal complement of $g(x)$.

\begin{figure}
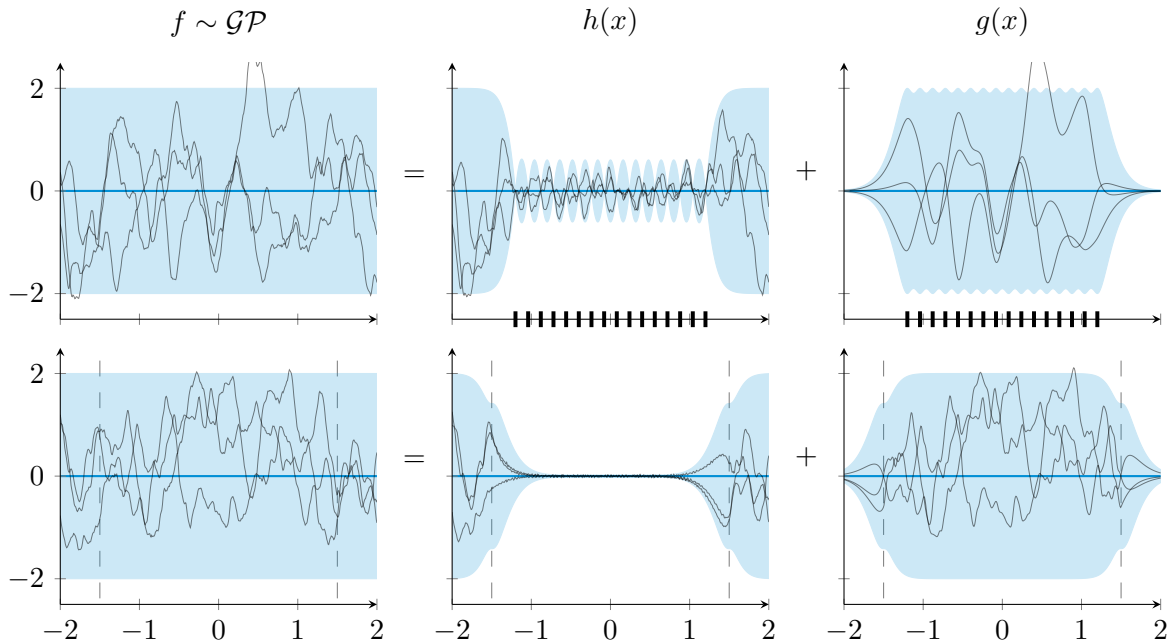

    \setlength\figurewidth{0.38\textwidth}
    \centering

     \caption{\label{fig:decomp} Decomposition of a Gaussian process with a \maternthreetwo covariance function ($\sigma=1, \ell=0.2$) using inducing points (top) and Variational Fourier Features (bottom). The shaded region represents two standard deviations, and samples from the processes are shown as black lines. In the top row, the positions of the 16 inducing points are marked in black; on the bottom row 31 frequencies (hence 63 inducing variables) are used and vertical lines denote the boundary of the projection $[a, b]$.
        Note the variance of $h(x)$ is zero wherever there is an inducing point (top). In the bottom row, the variance is close to zero for a continuous stretch (where the data should live); in this region $h(x)$ contains only high-frequency functions that we have failed to capture with the Fourier features. \add{The variance increases towards the boundaries: since all Fourier basis functions have the same value at $a$ and $b$, they cannot fit functions with different values in $a$ and $b$ and this is left as a residual. Instead of adding more basis functions, such as a linear one, to account for this, we advocate for choosing an interval $[a,b]$ larger than the data.}}
\end{figure}

Figure~\ref{fig:decomp} shows the decomposition of a Gaussian process using inducing points and also with the Fourier features that we shall derive in this document.

In the variational approximation, the posterior is controlled through the variables $\bu$, and the conditional $p(f(x)\given \bu)$ remains the same in the approximate posterior as in the prior. In terms of the projection, we see that the process $g(x)$ is completely controlled by the variables $\bu$ and the process $h(x)$ is completely {\em independent} of $\bu$. This is borne out by the independence relations of the processes, formally:
\begin{align}
  g(x) \given \bu &= \mathcal {GP}\left(\bk_\bu(x)^\top\Kuu^{-1}\bu, \,0\right)\,,\\
        h(x) \given \bu &= h(x)\,.
\end{align}
The assumption of the sparse variational framework is that the posterior process can be captured by the random variables $\bu$, or equivalently by the process $g(x)$. The power of the framework lies in the fact that the \textsc{ELBO} can be used to evaluate whether the projection is sufficient, and that the orthogonal part $h(x)$ is not discarded\remove{, but used in computation of the \textsc{ELBO}} \add{but taken into account in the \textsc{ELBO} computation} and incorporated fully at predict time. The \textsc{ELBO} encapsulates the degree to which the posterior can be captured by $g(x)$ (or equivalently $\bu$).

\subsection{Inter-domain approaches}
\label{par:inter}
The most common method for efficient approximation of Gaussian processes is based on the idea of inducing variables, denoted $u_m = f(z_m)$ in the above. A powerful idea is to generalize this, allowing a different decomposition of the process by constructing linear combinations of the process values as projections, instead of simple evaluations.

Previous work that has considered inter-domain approaches has suggested that the corresponding features have better ability to represent complicated functions \citep{figueiras2009inter} than the inducing points approach, and have promise for applications in multiple-output settings \citep{alvarez2009sparse}.

In inter-domain Gaussian process approximations, the idea is to change
\begin{equation}
    u_m = f(z_m)
\end{equation}
to a projection
\begin{equation}
    u_m = \int f(x) \dee\mu_m(x)\,,
\end{equation}
so that the variable $u_m$ is more informative about the process, and in turn the projection \remove{on to} \add{onto} $\bu$ is able to capture more information about the posterior.

The utility of the variable $u_m$ depends mostly on its covariance with the remainder of the process, which is encapsulated in the
vector-valued function $\bk_{\bu}(x)$. This vector plays a similar role in the variational method to the feature vector $\bphi(x)$ in the Fourier-based approximate models of Sections \ref{par:rff} and \ref{par:optff}. The remit of this work is to construct inducing variables $\bu$ such that $\bk_\bu(x)$ contains sinusoidal functions as is $\bphi(x)$, which we will do by using Fourier projections of the process.

The central challenge is that the Fourier transform of the process does not make a valid inducing variable because its variance would be infinite. To make valid and useful inducing variables, \citet{matthews2015sparse} have shown that the inducing variables $u_m$ must be ``{\em deterministic, conditioned on the whole latent function}''. In the following section, we construct valid inducing variables and examine their properties.
 \section{Variational Fourier Features}\label{par:vff}
Our goal is to combine the variational sparse GP idea with Fourier features, so
that the approximating process in \eqref{eq:gp_cond_sparse} contains a mean
function which is built from sinusoids, like the $\bphi(x)$ features in the
Fourier-features approximate model \eqref{eq:rff_model}.

This is more tricky than it first appears. An initial idea would be to define $u_m$ as the Fourier transform of the process, so that $\bk_\bu(x) = \cov(\bu, x) \propto \bphi(x)$. The question is, what would the $\Kuu$ matrix be? One might hope for a diagonal matrix related to the spectral content of the kernel $s(\omega)$, but
as we discuss in Section~\ref{par:trouble}, the Fourier transform of a stationary GP has diverging variance, i.e.\ this matrix is not defined.

\add{To proceed, we need to make two alterations to the initial idea, which we outline in Sections \ref{par:L2} and \ref{par:RKHS_FF}. First we must window the domain, in order to obtain variables with finite variance (Section \ref{par:L2}). Second, since these variables do not have an elegant form, we switch from an $L^2$ integral to one under the RKHS norm (Section \ref{par:RKHS_FF}).}
Subsequently, we show that our approach has covariance structures that make for efficient computations (Section~\ref{par:cov_structure}), and conclude this section by explaining the behaviour of our approximation outside the windowing box (Section~\ref{par:outside_box}).

\subsection{The trouble with Fourier features}
\label{par:trouble}
In the variational sparse GP (Section~\ref{par:sparseGP}) we made use of the random variables $u_m = f(z_m)$. To obtain a Fourier-based approximation, we might consider the inter-domain variables $u_m = \infint f(t) e^{-i \omega_m t} \dee t$. It is straightforward to show that such variables have zero mean and infinite variance, and so will not be useful in the variational framework; nonetheless the properties of the variables are interesting as we now demonstrate.

In order to use $u_m$ as an inducing variable, we would require both $\cov(u_m, f(x))$ and $\cov(u_m, u_{m'})$ which make up the entries in $\bk_{\bu}(x)$ and $\Kuu$ respectively. If $f$ is drawn from a GP with zero mean and covariance $k(x, x')$, then formally we may write
    \begin{equation}
    \cov(u_m, f(x)) = \mathbb E \left[ u_m f(x) \right] = \infint \mathbb E[f(t) f(x)] e^{-i \omega_m t} \dee t = \infint k(t, x) e^{-i \omega_m t} \dee t\,.
\end{equation}

To solve this integral, we can first plug in the definition of the kernel function in terms of the spectrum \eqref{eq:inv_fourier}, and then change the order of integration, giving
\begin{align}
    \cov(u_m, f(x)) &= \infint \infint s(\omega) e^{i \omega (t-x)} \dee \omega\, e^{-i\omega_m t}\dee t                            = s(\omega_m) e^{-i\omega_m x}\label{eq:inf_ff_cov} \,. \vphantom{\infint}
\end{align}
This initially appears promising: the covariance function is a sinusoid rescaled by the relevant spectrum and the elements of $\bk_\bu(x)$ have the desired form. The corresponding elements of $\Kuu$ are given by the (co)variance of $u_m$. It is here that problems arise. Denoting the complex conjugate of $u_m$ as $\bar u_m$,
\begin{align}
    \cov(u_m, u_{m'}) = \mathbb E\left[u_m \bar u_{m'}\right] =    \infint\infint k(t, t')e^{-i\omega_m t}\dee t\, e^{i\omega_{m'} t'}\dee t'
    &=s(\omega_m)\delta(\omega_m - \omega_{m'})\,,
\end{align}
where $\delta(\cdot)$ is Dirac's delta.
This implies that the matrix $\Kuu$ is diagonal, but with undefined entries on the diagonal. This result shows that it is simply not possible to use the Fourier transform of the whole GP as a valid inducing variable.

One approach to address this issue might be to take a power-averaged Fourier transform of the form
\begin{align}
    u_m = \lim_{a\to\infty} \frac{1}{a} \int_{-a/2}^{a/2} f(t) e^{-i\omega_m t} \dee t\,.
\end{align}
Unfortunately, this still does not yield useful features, because the variables $\bu_m$ are now independent of the function at any particular value. Writing formally:
\begin{align}
    \cov(u_m,f(x)) = \mathbb E[u_m f(x)] = \lim_{a\to\infty} \frac{1}{a} \int_{-a/2}^{a/2} k(x, t) e^{-i\omega_m r} \dee t = 0\,.
\end{align}
As recognised by \citet{figueiras2009inter}, the only way to obtain valid features is to specify an input density \citep[cf.][]{Williams+Seeger:2000}. We proceed by considering a uniform input density on the interval $[a, b]$.

\subsection{$L_2$ Fourier features on $[a, b]$}
\label{par:L2}
The reason that the Fourier transform of a GP behaves so strangely is that it
must explain the behaviour over the whole real line. One might interpret the
Fourier transform at a particular frequency as a sum of all the random
variables in the GP (i.e.\ integrating over the real line) multiplied by a
sinusoidal weight. The result is a Gaussian random variable, but unless the weight
decays to zero at infinity, the result has infinite variance.

Our solution to this diverging integral is to window the integral. \citet{figueiras2009inter} also use a window to ensure convergence of the integral, but their choice of a Gaussian windowing function means that the result is tractable only for Gaussian (squared exponential) covariance functions, and does not admit a diagonal form for $\Kuu$.

We will apply a square window, effectively changing the integration limits to $a$ and $b$: \sloppy
\begin{align}
    u_m = \int_{a}^{b} f(t) e^{-i \omega_m (t-a)} \dee t\,.
    \label{eq:l2_inner}
\end{align}
In addition, we will assume that the frequency $\omega_m$ is harmonic on the interval $[a, b]$; that is
\begin{equation}
    \omega_m = \frac{2 \pi m}{b-a}\,.
    \label{eq:harmonic_omegas}
\end{equation}
For $x \in [a,b]$, the covariance between such inducing variables and the GP at $x$ is
\begin{align}
    \cov(u_m, f(x)) = \mathbb E \left[ u_m f(x) \right] = \int_a^b \mathbb E[f(t) f(x)] e^{-i \omega_m (t-a)} \dee t = \int_a^b k(t, x) e^{-i \omega_m (t-a)} \dee t\,.
\end{align}

These integrals are tractable for \matern kernels. As detailed in Appendix~\ref{par:l2ff}, we obtain the following results for \maternonetwo kernels:
\begin{align}
    \cov(u_m, f(x)) = s_{1/2}(\omega_m) e^{i\omega_m (x-a)} + s_{1/2}(\omega_m) \frac{1}{2 \lambda}\left(\lambda[-e^{a-x} - e^{x-b}] + i\omega_m[e^{a-x} - e^{x-b}]\right)\,.
\end{align}
Similarly, the covariance between two inducing variables also has a closed form expression.
A complete derivation is given in Appendix~\ref{par:l2ff} for the \maternonetwo
case. These inter-domain inducing variables have two desirable properties: they
have finite variance and their associated covariance matrix $\Kuu$ can be written as a
diagonal matrix plus some rank one matrices (two in the \maternonetwo case).
\remove{Furthermore, the feature vector $\bk_\bu(x)$, which is made by evaluating
$\cov(u_m, f(x))$ is almost sinusoidal, aside from some rescaling and edge
effects.}

Figure~\ref{fig:edges} illustrates \add{entries of the feature vector $\bk_\bu(x) = \cov(u_m, f(x))$}, and compares \add{them} to the RKHS features that we shall derive in the next section. We see from the Figure that the covariance function $\cov(u_m, f(x))$ is \add{almost} sinusoidal in $x$ for a region sufficiently far from the boundaries \remove{$[a, b]$} \add{$a$ and $b$}. The `edge effects' depend on the involved frequency as well as the lengthscale parameter of the kernel.

\add{We have constructed inducing variables that are valid (in the sense that they have finite variance), but the result is somewhat inelegant: the expressions for $\cov(u_m, f(x)$ and $\cov(u_m, u_{m'})$ are long for the \maternonetwo kernel and become trickier (but still tractable) for higher orders.} \remove{Whilst these inducing variables are valid in the sense that they have finite variance, w}We abandon them at this stage in preference of the RKHS-based inducing variables that we derive in the next section. These have preferable edge effects and an elegant expressions for the required covariances.

\begin{figure}
    \setlength\figurewidth{0.49\textwidth}
    \setlength\figureheight{0.3\textwidth}
\begin{center}
 \end{center}
    \caption{\label{fig:edges} The covariance $\cov(u_m, f(x))$ as a function of $x$ for the \maternonetwo (top), \maternthreetwo (middle) and \maternfivetwo (bottom) kernels, using $L_2$ Fourier features (left column) and RKHS Fourier features (right column). The red and blue lines represent the real and imaginary parts of the covariance. The boundaries of the projection $[a, b]$ are marked by vertical dashed lines. The $L_2$ features are \remove{sinusoidal for a small central section far from the boundary} \add{wiggly with overwhelming edge effects}, whereas the RKHS features are exactly sinusoids for $x \in [a,b]$. Beyond the \remove{boundary} \add{boundaries}, the covariance reverts to zero in a way which depends on the smoothness of the kernel and the lengthscale parameter. We have used $\omega_m=16\pi/(b-a)$, and a lengthscale of $\ell=(b-a)/10$.}
\end{figure}

\subsection{RKHS Fourier features on $[a, b]$}
\label{par:RKHS_FF}
In the previous section we saw that choosing inducing variables of the form $u_i = \PSi{f,\cos(\omega (x - a)}{L^2}$ or $u_i = \PSi{f,\sin(\omega (x - a)}{L^2}$ resulted in features that are very distinct from cosine and sine functions.  In this section, \remove{we will use some RKHS theory to construct the decomposition directly , and then show that corresponding random variables $\bu$ exist.} \add{we replace the $L^2$ inner product by the RKHS inner product $u_i = \PSi{f,\cos(\omega (x - a)}{\mathcal H}$ in order to guarantee that the features are exactly the sine and cosine functions. As we will discuss later, a direct asset of this approach is that the simple expression of the features makes the computation of the covariances between inducing variables much easier.}

We start with a truncated Fourier basis defined similarly to \eqref{eq:basis_expansion}
\begin{align}
    \bphi(x) = [1, \cos(\omega_1 (x-a)), \ldots, \cos(\omega_M (x-a)), \sin(\omega_1 (x-a)),\ldots,\sin(\omega_M (x-a))]^\top\,,
\end{align}
where we include the constant basis function, $\phi_0(x) = 1$, accounting for $\omega_m=0$, and define $\omega_m = \frac{2\pi m}{b-a}$ as previously.

A key RKHS result~\citep[see][Theorem 11]{berlinet2004reproducing}, is that if $\cF = \textrm{span}(\bphi)$ is a subspace of a RKHS $\cH$, then $\cF$ has the kernel
\begin{align}
k_\cF(x,x') = \bphi(x)^\top \Kpp^{-1} \bphi(x')\,,
\end{align}
where $\Kpp[m,m'] = \PS{\phi_m,\phi_{m'}}$ is the Gram matrix of $\bphi$ in $\cH$. By this, we mean that for any function $f\in\cF$, $\PS{f, k_\cF(\cdot, x)} = f(x)$. Furthermore, the coordinate of the projection of a function $h \in \cH$ onto $\phi_m$ is defined as
\begin{align}
\label{eq:projRKHS}
    \cP_{\phi_m} (h) = \PS{h,\phi_m} \, .
\end{align}

Since $\cF \subset \cH$, $k(x, x') - k_\cF(x, x')$ is a positive definite function corresponding to the kernel of $\cF^\perp$, and we can decompose the GP in a similar way to \eqref{eq:gp_project1} which gives
\begin{align}
    g(x) &\sim \cGP(0,\, \bphi(x)^\top\Kpp^{-1}\bphi(x'))\,,\\
    h(x) &\sim \cGP(0,\, k(x, x') - \bphi(x)^\top\Kpp^{-1}\bphi(x'))\,,\\
    f(x) &= g(x) + h(x)\,.
\end{align}

\add{
For this to be valid we need two things: $\cF$ must be included in $\cH$ and the matrix $\Kpp$ must be invertible. The first is not true in general (\matern RKHS over $\mathbb{R}$ or the RBF and Brownian RKHS on $[0,1]$ are counter examples) but has been proved to be true for \matern kernels over $[a,b]$ \citep{durrande2016detecting}. The second is a direct conclusion of the linear independence between the elements of $\bphi$. Furthermore, \citet{durrande2016detecting} also detail the closed-form expressions of the inner products for the \maternonetwo, \maternthreetwo and \maternfivetwo RKHSs. These expressions, which are repeated in Appendix~\ref{par:matern_inners}, are of particular interest here since they allow us to compute the entries of the $\Kpp$ matrix.  Although we refer to the above reference for the constructive proof of these expressions, it is easy to check that the reproducing property is satisfied. For example, the expression given in the appendix for a \maternonetwo RKHS over $[a,b]$ is
\begin{align}
\PSi{g,h}{\cH_{1/2}} & = \frac{1}{2 \lambda \sigma^2} \int_a^b \left( \lambda g(t) + g'(t) \right) \left( \lambda h(t) + h'(t) \right) \dee t + \frac{1}{\sigma^2} g(a)\,h(a)\,,
\label{eq:PSk12}
\end{align}
so we can write for any function $g$ that is continuous and (weakly) differentiable
\begin{align}
\PSi{g,k_{\tfrac{1}{2}}(x,\cdot)}{\cH_{1/2}} & = \frac{1}{2 \lambda \sigma^2}\left( \int_a^x \left( \lambda g(t) + g'(t) \right) 2 \lambda \sigma^2 e^{\lambda (t-x)} \dee t + \int_x^b 0 \dee t \right) + g(a) e^{\lambda (a-x)}\\
& = \int_a^x \lambda g(t)  e^{\lambda (t-x)} \dee t + \left[ g(t)  e^{\lambda (t-x)} \right]_a^x - \int_a^x  g(t) \lambda e^{\lambda (t-x)} \dee t + g(a) e^{\lambda (a-x)}\\
& = g(x) - g(a) e^{\lambda (a-x)} + g(a) e^{\lambda (a-x)}\\
& = g(x) \,.
\label{eq:PSk12proof}
\end{align}
Another property of this inner product is that the Gram matrix $\Kpp$ has a particular structure which allows to reduce drastically the computational burden of computing and inverting $\Kpp$. This will be addressed in Section~\ref{par:cov_structure}.
}
\subsubsection{The corresponding random variables}
To interpret these results from a Gaussian process point of view, we need to find the inducing variables $\bu$ that would lead to the exact same decomposition.

We would like to use the RKHS inner product between the sinusoids and the GP sample path in place of the $L_2$ inner product \eqref{eq:l2_inner}, but
since GP samples do not belong the RKHS \citep{driscoll1973reproducing}, it is \emph{a priori} not possible to apply $\cP_{\phi_m}$ to $f$ and define $u_m = \PS{\phi_m,f}$. For example, the \maternonetwo inner product involves derivatives whereas the \maternonetwo samples are not differentiable anywhere. However, using the fact that the functions from $\bphi$ are very regular, it is possible to extend the operators $\cP_{\phi_m}: h \mapsto \PS{\phi_m,h}$ to square integrable functions using integration by parts. For the \maternonetwo kernel, integrating \eqref{eq:PSk12} results in
\begin{equation}
\begin{split}
    \cP_{\phi_m}(h) & = \frac{1}{2 \lambda \sigma^2} \left( \int_a^b h ( \lambda^2  \phi_m - \phi_m'' ) \dee t + h(b) \left( \lambda  \phi_m(b) + \phi_m'(b)\right) + h(a) \left( \lambda  \phi_m(a) - \phi_m'(a) \right)  \right) \,.
    \label{eq:integrated_inner}
 \end{split}
\end{equation}
Note that the function $h$ does not need to be differentiated, and so it is possible to apply this functional to a sample from the GP. Similar results apply for the \maternthreetwo and \maternfivetwo kernels. It is now possible to apply these operators to the Gaussian process in order to construct the inducing variables,
\begin{align}
    u_m = \cP_{\phi_m}(f)\,.
\end{align}

The covariance of the inducing variables with the function values is given by
\begin{align}
    \cov(u_m, f(x)) = \mathbb E[u_m f(x)] = \mathbb E[\cP_{\phi_m}(f) f(x)] = \cP_{\phi_m}\left(k(x, \cdot)\right) = \phi_m(x)
\end{align}
which is valid for $a \leq x \leq b$. This means that $\bk_\bu(x) = \bphi(x)$ when $x\in [a, b]$.  Similarly the covariance of the features is given by
\begin{align}
    \cov(u_m, u_{m'}) = \mathbb E[u_m u_{m'}] = \mathbb E[\cP_{\phi_m}(f) \cP_{\phi_{m'}}(f)] = \cP_{\phi_m}(\phi_{m'}) = \langle\phi_m,\phi_{m'}\rangle_\cH\,,
\end{align}
and so we have $\Kuu = \Kpp$.

\subsection{Covariance structures for RKHS Fourier features}
\label{par:cov_structure}
To compute the covariance matrix $\Kuu$, we return again to the closed-form expressions for the inner product provided by \citet{durrande2016detecting}, substituting the basis functions $\phi_m$ and $\phi_{m'}$ appropriately. The solutions are tractable, \add{and we detail in the appendix the expressions of $\Kuu$ for the first three half-integer \matern kernels. One striking property of the resulting $\Kuu$ is that they can be written as a diagonal matrix plus a few rank one matrices. For example in the \maternonetwo case} \remove{for the \maternonetwo kernel} we find that $\Kuu$ is equal to a diagonal matrix plus a rank one matrix,
\begin{equation}
\label{eq:Kuu_dec_rank}
    \Kuu = \textrm{diag}(\balpha) + \bbeta\bbeta^\top\,,
\end{equation}
with
\begin{alignat}{21}
    &\balpha&&=\frac{b-a}{2}&&\big[&&2 s(0)^{-1}&&,\, &&s(\omega_1)^{-1}&&,\, &&\ldots\,&&,\, &&s(\omega_M)^{-1}&&,\, &&s(\omega_1)^{-1}&&,\, &&\ldots\,&&,\, &&s(\omega_M)^{-1}&&\big]^\top\,,\\
    &\bbeta&&=                       &&\big[&&\sigma^{-1}&&,   &&\,\sigma^{-1}                  &&, &&\ldots\,&&,  &&\,\sigma^{-1}                  &&, &&\,0                       &&, &&\ldots\,&&,  &&\,0&&\big]^\top\,.
\end{alignat}
        \remove{The expressions of $\Kuu$ for the \maternthreetwo and \maternfivetwo RKHSs are detailed in Appendix~\ref{par:matern_inners}. For such kernels,} \add{As shown in the appendix,} $\Kuu$ still has a similar structure \add{for higher order \matern kernels}: in the \maternthreetwo case, it is the sum of a diagonal matrix and two rank one matrices and in the \maternfivetwo case, it is the sum of a diagonal matrix and three rank one matrices.

Since $\Kuu$ has a low-rank plus diagonal form, one of the usual computational bottlenecks in the variational framework, solving $\Kuu^{-1}\Kuf$, can be done in $\mathcal O(NM)$ operations, rather than the standard $\mathcal O(NM^2)$, by a straightforward application of the Woodbury matrix identity.

This low-rank plus diagonal structure was not published previously in \citet{durrande2016detecting}; we believe that we are the first to make use of this result for computational efficiency and we expect it to provide similar superior computational benefits as the pure diagonal structure in \citet{solin2014hilbert}.

\subsection{Predicting outside the interval $[a, b]$}
\label{par:outside_box}
In the preceding sections, we have defined our inducing variables and how they covary with the function; we have assumed throughout that we wish to examine a point on the latent function in the pre-defined interval $[a, b]$. This does not pose a problem at training time, since we can easily define $[a,b]$ to contain all the data. For prediction we may need to predict outside this interval. For a point on the function $f(x)$ outside the interval $[a, b]$, the covariance with the inducing variables is still well defined, though the form is slightly more complicated than the simple sinusoids inside the interval.

To obtain the expressions for the covariance beyond $[a, b]$, we apply the closed-form expression for the covariance \eqref{eq:integrated_inner}, this time for $\cov(u_m, f(x)) = \cov(\cP_{\phi_m}(f), f(x))$ with $x > b$ and $x <a$. After some fairly elementary simplifications, the results are shown in Table~\ref{tab:extensions}, and illustrated in Figure~\ref{fig:edges}. The result is that the covariance function $\cov(u_m, f(x))$ returns to zero beyond the boundary, with smoothness that depends on the order of the kernel.

\begin{table}
    \begin{center}
    \begin{tabular}{c c c c} \toprule
        $\phi_m(x),\,x\in [a, b]$&\maternonetwo&\maternthreetwo&\maternfivetwo\\ \midrule
        $\cos(\omega_m(x-a))$
        & $e^{-\lambda r}$
        & $(1 + \lambda r) e^{-\lambda r}$
        & $(1 + \lambda r + \tfrac{1}{2}(\lambda^2 - \omega_m^2)r^2)e^{-\lambda r}$\\
        $\sin(\omega_m(x-a))$
        & 0
        & $sr\omega_me^{-\lambda r}$
        & $sr\omega_m(1 + \lambda r) e^{-\lambda r}$\\ \bottomrule
    \end{tabular}
    \end{center}
    \caption{The covariance $\cov(u_m, f(x))$ for $x$ outside the interval $[a, b]$. Here, we define $r$ as the absolute distance to the closest edge ($a$ or $b$), and $s$ to be $1$ for $x < a$ and $-1$ for $x > b$.\label{tab:extensions}}
\end{table}
 \section{Extending the applicable kernel family}\label{par:extending}
In the above we have focused on the \matern family of kernels in one dimension. In this section we will expand the approach to higher-dimensional inputs using sums and products of kernels.

\subsection{Additive kernels}
\label{par:additive_kernels}
A straightforward way to produce a Gaussian process prior on multiple inputs (say $D$) is to use a sum of independent Gaussian processes, one for each input:
\begin{equation}
    f(\bx) = \sum_{d=1}^D f_d(x_d)\,,\qquad f_d \sim \cGP\left(0,\, k_d(x_d, x_d')\right)\,,
\end{equation}
where $x_d$ is the $d$th element in $\bx$ and $k_d(\cdot,\cdot)$ is a kernel defined on a scalar input space. This construction leads to a Gaussian process with an additive kernel, that is we can write
\begin{equation}
    f(\bx) \sim \cGP\left(0,\, \sum_{d=1}^D k_d(x_d, x_d')\right)\,.
\end{equation}
Additive kernel structures have been explored by \citet{durrande2011additive} and \citet{duvenaud2011additive}, who have shown that these kernels are well suited to high-dimensional problems. To use our Variational Fourier Features approach with an additive kernel, we assume that each function $f_d(x_d)$ has a \matern covariance function, and decompose each of these GPs. We construct a matrix of features, with elements
\begin{equation}
    u_{m,d} = \cP_{\phi_m}(f_d)\,.
\end{equation}
The result is that we have $DM$ features. By construction, we see that features from different GPs are independent, $\cov(u_{m, d}, u_{m, d'}) = 0$, and that the covariance between features for the same dimension follows the construction for a single dimension. It is straightforward to exploit these independence structures for computational scalability during inference.

\subsection{Separable kernels}
\label{par:kron}
A kernel is said to be separable if it can be written as a product of kernels with no shared inputs. This means that for any $D$-dimensional input $\bx$, we can write
\begin{equation}
    k(\bx,\bx') = \prod_{d=1}^D k_d(x_d,x_d')\,.
\end{equation}
We can construct Variational Fourier Features for such a product kernel if each sub-kernel is in the \matern family.

This formulation of separability has been extensively used in speeding up GP inference \citep[see, e.g.,][]{Saatchi:2012, stegle2011efficient} by writing the covariance matrix as a Kronecker product of the individual covariance matrices. These speed-ups usually require the data to be aligned with a regular or rectilinear grids \citep[see, e.g.,][]{solin2016regularizing}, though some approaches exist to extend to observations beyond the grid \citep{wilson2014fast, nickson2015blitzkriging}, requiring additional approximations.

In contrast, our approach of decomposing the kernel naturally leads to a Kronecker structure, even for irregularly spaced data. Let $f(x) \sim \cGP\bigg(0, \prod_d k_d(x_d, x_d')\bigg)$, and define a vector of features as the Kronecker product of features over each dimension,
\begin{align}
    \bphi(\bx) = \bigotimes_{d=1}^D \left[\phi_1(x_d),\ldots,\phi_M(x_d)\right]^\top\,,
\end{align}
so that each of the $M^D$ elements of $\bphi(\bx)$ is a product of one-dimensional functions $\prod_d \phi_{i}^{(d)}(x_d)$. We define a hyper-rectangular boundary given by $\prod_d [a_d, b_d]$, and define the inducing variables $\bu$ using a projection similar to the one-dimensional case
\begin{align}
    u_m = \cP_{\phi_m}(f)\,,
\end{align}
where we use the norm of the Hilbert space associated with the product kernel.
The covariance between an inducing point $u_m$ and the function is then given by
\begin{align}
    \cov(u_m, f(\bx)) = \mathbb E[\cP_{\phi_m}(f) f(x)] = \cP_{\phi_m}(k(\bx, \cdot)) = \prod_{d=1}^D\phi_m^{(d)}(x_d) = \bphi_m(\bx)\,.
\end{align}
Extending the covariance function beyond the boundary follows similarly to the one-dimensional case above.
The covariance between inducing variables is given by
\begin{align}
    \cov(u_m, u_m') = \prod_d \PSi{\phi_m^{(d)}, \phi_{m'}^{(d)}}{\cH_d}\,.
\end{align}
This means that the covariance matrix $\Kuu$ has a Kronecker structure,
$\Kuu = \bigotimes_{d=1}^D\Kuu^{(d)}$, where each sub-matrix $\Kuu^{(d)}$ has the same structure as for the one-dimensional case \eqref{eq:Kuu_dec_rank}.
 \section{Implementation details and computational complexity}\label{par:details}
In this section we give some details of our implementation of the methods proposed, and consider the theoretical computational complexity of each. To make use of the RKHS Fourier features that we have derived, the expressions given for $\Kuu$ and $\bk_\bu(x)$ can simply be plugged in to the variational framework described in Section~\ref{par:sparseGP}.

\subsection{Implementation}
All of the methods in this paper and all of the code to replicate the experiments in Section \ref{par:experiments} are available at \url{http://github.com/jameshensman/VFF}. We made use of TensorFlow \citep{tensorflow} and GPflow \citep{matthews2017gpflow} to construct model classes in Python. Some simple matrix classes in Python assisted with computing efficient solutions to low-rank matrix problems (e.g.\ solving $\Kuu^{-1}\bm$). Using TensorFlow's automatic differentiation made application of gradient based optimization straightforward.

\subsection{The one-dimensional case}
Consider first problems with only one input dimension. The sparse variational framework discussed in Section~\ref{par:sparseGP} gives three methods: if the likelihood is Gaussian, then it is possible to compute the optimal posterior in closed form \citep[as per][]{titsias2009variational}, if the likelihood is not Gaussian we can approximate the posterior with a Gaussian or use MCMC on the optimal distribution.

For the Gaussian case, we see that the computation of the optimal posterior (equation~\eqref{eq:optimal_q}) is dominated by the matrix multiplication $\Kuf\Kfu$, which costs $\cO(NM^2)$ operations. In our case, since $\Kfu$ does not depend on the kernel parameters, this multiplication only needs to be computed once before optimizing (or sampling) the covariance function parameters. To compute the posterior mean and covariance, we must invert an $M\times M$ matrix, so the cost per iteration is $\cO(M^3)$, after the initial cost of $\cO(NM^2)$.

For the non-Gaussian case with a Gaussian approximation we must compute the marginals of $q(\bff)$, in order to evaluate the ELBO (equation~\eqref{eq:gp_elbo}) for optimization. We parameterize the Gaussian covariance using a lower-triangular matrix $\bL$, so $\bSigma = \bL\bL^\top$, and the computational cost is dominated by the matrix multiplication $\Kfu \Kuu^{-1} \bL$. Even though the matrix $\Kuu$ has computationally-convenient structure, we cannot avoid a dense matrix multiplication in this case. It may be possible that by assuming some structure for the approximate covariance, computation could be reduced, but we leave investigation of that line for future research.

We note however that the Gaussian approximation does lend itself to stochastic optimization, where the data are randomly sub-sampled \citep{hensman2013gaussian}. In this case the computational complexity is $\cO(\tilde N M^2)$, where $\tilde N$ is the size of the minibatch.

For the MCMC approach, evaluating the unnormalised distribution~\eqref{eq:q_hat} and its derivative costs $\cO(NM)$ per iteration, since again the  cost is dominated by the computation of the marginals of $q(\bff\given \bu)$, which requires only the product $\Kfu\Kuu^{-1}$, and the low-rank structure of $\Kuu$ enables this to be computed efficiently.

\subsection{The additive case}
If we consider the kernel to be a sum of kernels over $D$ input dimensions (Section~\ref{par:additive_kernels}), then we again have access to the three methods provided by the sparse GP framework. In this case, the number of inducing variables is $2MD+1$, since we have $M$ frequencies per dimension, with sine and cosine terms.

For the method with a Gaussian likelihood, the cost is $\cO(D^3M^3)$ per iteration, with an initial cost of $\cO(NM^2D^2)$, following the arguments above. When the likelihood is non-Gaussian, with a Gaussian approximation to the posterior, the cost is $\cO(NM^2D^2)$. However, if we assume that the posterior factorizes over the dimensions, then the cost can be reduced to $\cO(NM^2D)$ \citep{adam2016scalable}.

For an MCMC approach, the cost is again dominated by the computation of $\Kfu\Kuu^{-1}$, which must be done once for each dimension and so the cost is $\cO(NMD)$. We note that these operations can be effectively distributed on multiple machines (in our implementation this can be done by TensorFlow), since $\Kuu$ is a block-diagonal matrix.

\subsection{The Kronecker case}
If we use a separable kernel made from a product of kernels over the $D$ dimensions as discussed in Section~\ref{par:kron}, then the number of inducing variables grows exponentially with the input dimension to $(2M)^D$. This gives the potential for a very detailed representation, with many basis functions, and the Kronecker structure can be exploited for efficiency in some cases. However, this approximation will be unsuitable for large $D$.

For the case with Gaussian noise, it is not straightforward to avoid computing the inverse of the $M^D \times M^D$ matrix, and so the cost is $\cO(M^{3D})$. It may be possible, however, that some efficient methods for solving linear systems might be applicable \citep[e.g.][]{filippone2015enabling}, though we leave that avenue for future work.

For the case with a Gaussian approximation to $q(\bu)$, we can exploit the Kronecker structure of the model. We note the relationship to the work of \citet{nickson2015blitzkriging}, which used the inducing-point approach, with the points confined to a regular lattice. In their work, the covariance structure of the approximation was constrained to also have a Kronecker structure, which appeared to work well in the cases considered. However, our experiments suggest (Section~\ref{par:experiment_classification}) that this approximation may not be sufficient in all cases. Our proposal is to use a sum of two Kronecker structured matrices:
\begin{align}
     q(\bu) = \cN\left(\bm,\,\bS\right),\qquad \text{where }\bS= \bigotimes_{d=1}^D \bL_d\bL_d^\top + \bigotimes_{d=1}^D \bJ_d\bJ_d^\top\,.
\end{align}
Noting that we can compute the determinant of such a matrix structure in $\cO(DM^3)$ \citep[see e.g.][]{rakitsch2013all}, the computational cost of the method is then dominated by the cost of computing the mean and marginal variances of $q(\bff)$. For the mean, we first compute $\Kuu^{-1}\bm$ which costs $\cO(M^D)$ (making use of the low-rank structure of $\Kuu$), and then $\Kfu(\Kuu^{-1}\bm)$, which costs $\cO(NM^D)$.  In computing the marginal variances of $q(\bff)$, the additive structure of $\bS$ poses no problem, and the Kronecker structures mean that $\Kfu\Kuu^{-1}\bS$ can be computed in $\cO(NM^{2D})$. Again, it is possible to make use of stochastic optimization to reduce the cost per iteration.

In the MCMC case, the cost is once again dominated by computing the marginals of $q(\bff\given\bu)$, which costs $\cO(NM^D)$.

\subsection{\add{Centring} of variables for MCMC}
\label{par:rot_mcmc}
For an effective MCMC scheme, it is necessary to re-\add{centre} the variables $\bu$ using a square-root of $\Kuu$ \citep{christensen2006robust, vanhatalo2007sparse, murray2010slice, filippone2013comparative, hensman2015mcmc} such that
\begin{align}
    \bv &\sim \cN(\bzero, \bI)\,,\\
    \bu &= \Kuu^{1/2} \bv\,,
\end{align}
which gives $\bu \sim \cN(\bzero, \Kuu)$. For dense $\Kuu$ matrices, the square root $\Kuu^{1/2}$ is often chosen to be the Cholesky factor, but the structured $\Kuu$ described above has a more convenient square root, given by concatenating the two parts into
\begin{align}
\bR = \left[\textrm{diag}(\balpha)^{\tfrac{1}{2}},\,\bbeta \right]
\end{align}
so that $\bR\bR^\top = \textrm{diag}(\balpha) + \bbeta\bbeta^\top = \Kuu$. Since $\bR$ has an extra column (or two or three extra columns for \maternthreetwo and \maternfivetwo respectively) we require an extra variable(s) in the vector $\bv$.
 \section{Experiments}\label{par:experiments}
This Section provides empirical evidence and examples of how the Variational Fourier Features method works in practice. We first cover some illustrative toy examples, which underline \remove{aspects} \add{the} accuracy of the approximation and the influence of the approximation parameters. After this we address three standard test examples for GP regression (for high- and low-dimensional inputs) and classification, where we show the method \add{provides accurate predictions,  and does so extremely fast.}  Finally, we examine how the approximation approaches the posterior in the MCMC case by examining a simple log Gaussian Cox process dataset.

\subsection{Illustrative examples}
Our Variational Fourier Features (VFF) method has two tuning parameters which affect the approximation: the number of Fourier features $M$, and the choice of the bounds \remove{$[a,b]$} \add{$a$ and $b$}. In this Section the effects related to tuning these parameters are demonstrated.

\subsubsection{Regression}
\begin{figure}[t]
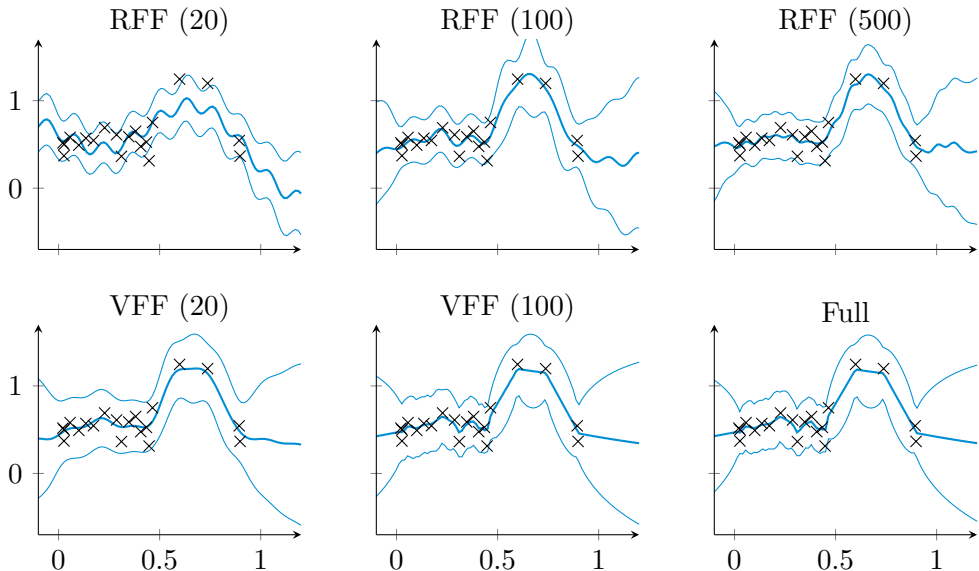

\centering
 \caption{\label{fig:rff_compare}
    A comparison of our Variational Fourier Features (VFF) with Random Fourier Features (RFF). Top row: Random Fourier Features with 20, 100, and 500 frequencies. Bottom row: Variational Fourier Features with 20 and 100 frequencies, and the full-GP fit. The kernel is a \maternonetwo, with variance, lengthscale and noise variance set by maximum likelihood according to the full GP. The bounds of the problem are set to $[-1, 2]$ for the VFF.}
\end{figure}

To perform GP regression with Variational Fourier Features, one simply takes the expressions given for standard GP approximations (equations~\eqref{eq:gp_pred_sparse} and~\eqref{eq:optimal_q}) and substitutes $\Kuu,\ \Kuf$ and $\bk_{\bu}(x)$ appropriately. Figure~\ref{fig:rff_compare} compares the VFF approximation with Random Fourier Features on a trivial but illuminating regression experiment. 

We started by fitting a `full' GP model (i.e.\ with inversion of the dense covariance matrix), maximising the likelihood with respect to the kernel's variance parameter, lengthscale parameter and the noise variance. We then used these fitted values to compare variational and Random Fourier Features.

The Random Fourier Features method has some attractive convergence properties \citep{rahimi2007random}, as well as being simple to implement. Here we show empirically that the VFF method approaches the true posterior much faster, in terms of the number of basis functions used.

This experiment uses a \maternonetwo kernel. The corresponding spectral density is proportional to a Cauchy distribution, which has heavy tails. The frequencies in Random Fourier Features are drawn from this heavy-tailed distribution. We see from Figure~\ref{fig:rff_compare} that this leads to a poor Monte Carlo estimate of equation~\eqref{eq:inv_fourier}. With 20 random features, the fit of the model is poor, whilst with 20 variational features, the approximation is qualitatively better. With 100 features, the random features approach still cannot replicate the full GP, whilst the variational method is \remove{almost} \add{already} exact \remove{. We observed that the ELBO was tight to the likelihood to within numerical precision} \add{since the ELBO is tight to the likelihood to within numerical precision}. Even with 500 features, the RFF method cannot approximate the model well.

This demonstration is exacerbated by the choice of kernel function. For other kernels in the \matern series, with spectra that behave more like a Gaussian, the problem is less severe. Nonetheless, this demonstrates the power of Variational Fourier Features for these models.

\subsubsection{Setting $a$, $b$, and $M$}
\begin{figure}
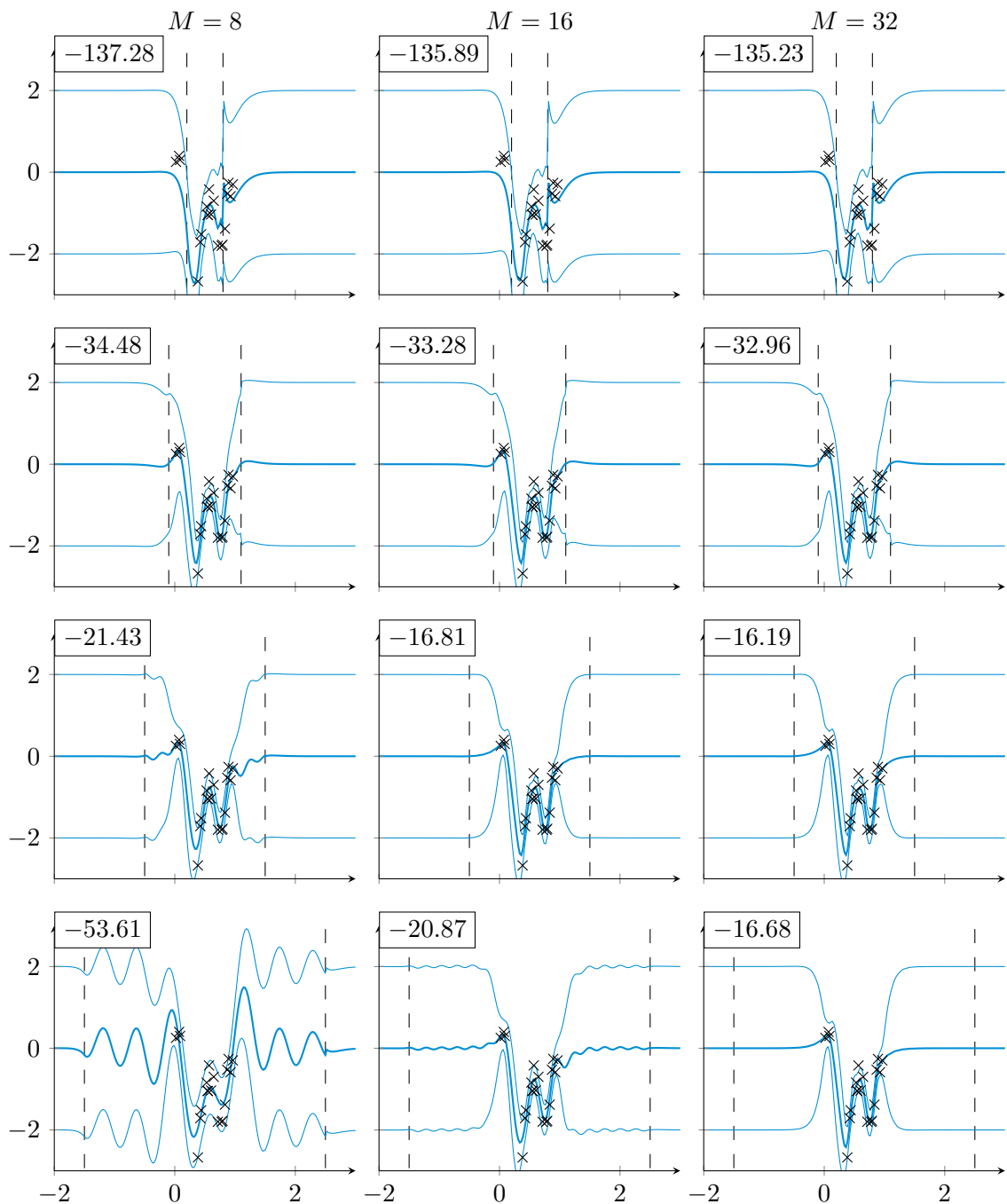

    \setlength\figurewidth{0.4\textwidth}
    \centering\pgfplotsset{every axis title/.append style={below right, at={(0,1)}, draw=black, fill=white}}

     \caption{\label{fig:choosing_interval}Demonstration of the effect of changing the interval size $[a, b]$ and the number of frequencies used. Each column has an increasing number of frequencies ($M=8, M=16, M=32$), and each row has an increasing interval size, marked by vertical dashed lines. The data are drawn from a \maternthreetwo kernel with Gaussian noise, and inference is performed with the kernel parameters set to the known values. The ELBO is shown in the top left of each plot. For reference, the true marginal likelihood in this problem is $-15.99$. }
\end{figure}

To use Variational Fourier Features, we must select the \remove{boundary of the} projection \add{interval} $[a, b]$, as well as the frequencies used. In all our experiments, we use the first $M$ frequencies that are harmonic on the interval, that is $\mathbf \omega = [\frac{2\pi m}{(b-a)}]_{m=1}^M$. 
Figure~\ref{fig:choosing_interval} compares the effect of changing the interval size and the number of inducing frequencies, for a simple Gaussian regression problem. We drew data from a Gaussian process with a \maternthreetwo kernel ($\sigma^2=1,\, \ell=0.2$) and added Gaussian noise ($\sigma^2_n=0.05$). We used these known parameters for the series of regression fits shown, varying $[a, b]$ and $M$. The true marginal likelihood was computed to be $-15.99$.

The top row has the interval set too narrow, within the region of the training data (this is valid but not recommended!). Accordingly, the ELBO (marked in the upper left corner of each plot) is low, no matter how many frequencies are used. In the next row, the boundary is still too close to the data: some edge effects are visible. The low value of the ELBO reflects the low quality oif the approximation. 

In the third row, the boundary $[a, b]$ is sufficiently far from the data. No edge effects are visible, but we notice extra wiggles in the approximation when insufficient frequencies are used. For sufficient frequencies, and this sensible setting of the interval, the ELBO is close to the marginal likelihood.

The fourth row of Figure~\ref{fig:choosing_interval} shows the effect of choosing an interval that is too large. Because we have set the frequencies to be dependent on $(b-a)$, making this quantity large makes the frequencies too low. Sinusoidal behaviour is visible with $M=8$ or $M=16$, although a sensible solution is recovered once $M=32$. Again, the ELBO reflects the quality of the approximation.

The purpose of this experiment was to illustrate how the properties of the approximation parameters ($a, b,$ and $M$) affect the approximation. We have emphasized how the ELBO can be examined to ensure that two key considerations are met. First, that $a$ and $b$ are sufficiently far from the edges of the data. Second, that a sufficient number of frequencies are used. We emphasize again the monotonic nature of the approximation inherited from the variational framework: adding more features (increasing $M$) necessarily improves the approximation in the KL sense.

\subsubsection{Comparison with Sparse GP for various input dimensions}

The computational complexity of the VFF approach increases with the dimension of the input space. For additive kernels, we have shown that the complexity increases linearly with input dimension, and for models with a product kernel, the computation increases exponentially. In this section we explore the quality of our approximation in the product-kernel case, and compare with the sparse (inducing input) approach. We measure the time/accuracy tradeoff as a function of the input dimension and the number of inducing variables.

We consider toy datasets with input dimension $d=1\ldots 4$, with input locations drawn uniformly over $[0,1]^d$. The response data are given by GP samples with products of \maternthreetwo kernels, with additive Gaussian observation noise:
\begin{align}
	x_{i,j} & \sim \mathcal{U}(0,1)&& \text{for } 1 \leq i \leq 10^4 \text{ and } 1 \leq j \leq d \, ,\nonumber\\
	\bff & \sim \cN\bigg(\bzero, \prod_{j=1}^d k_{3/2}(\bX_j,\bX_j)\bigg) \, ,&&\\
	\by & \sim \cN\big(\bff,\, 0.1 \times \bI\big) \, .&&\nonumber
\end{align} 

Given these data, we compare two approximations: one sparse GP with $M=m^d$ inducing inputs located on a regular grid with minimal and maximal values equal to 0 and 1 for each coordinate and a VFF approximation based on a Kronecker structure with $M=(m-1)/2$ frequencies for each input dimension which also leads to $m^d$ inducing variables. For both approximations, the parameters of the covariance function were fixed to the known values used to generate the data (lengthscales $\ell=0.2$, variance $\sigma^2=1$). The domain boundaries $[a,b]$ that are specific to the VFF model were fixed to $[-0.3,1.3]$, which seemed to be a reasonable trade-off for the various dimensions after testing a few other values. We then compute the KL divergence between each approximation and the truth, by comparing the ELBO to the marginal likelihood from a `full' GP regression model. We recorded the time required to compute the ELBO, and repeated the whole experiment five times.

\begin{figure}
  \captionsetup[subfigure]{aboveskip=-0.3ex, belowskip=2ex}
  \centering
  \setlength\figurewidth{0.47\textwidth}
  \setlength\figureheight{0.85\figurewidth}
      \pgfplotsset{yticklabel style={text width=1cm, align=right}}
    \begin{subfigure}[t]{0.49\textwidth}
        \begin{tikzpicture}

\definecolor{color0}{rgb}{0.12156862745098,0.466666666666667,0.705882352941177}
\definecolor{color1}{rgb}{1,0.498039215686275,0.0549019607843137}

\begin{axis}[
xlabel={time (s)},
ylabel={$\textsc{KL}\big[q\big(f(x)\big) || p\big(f(x)\given \by\big)\big]$},
xmin=0.09, xmax=0.19,
ymin=-907.973408424743, ymax=19319.5805507148,
xmode=log,
width=\figurewidth,
height=\figureheight,
tick align=outside,
tick pos=left,
x grid style={white!69.019607843137251!black},
y grid style={white!69.019607843137251!black},
legend cell align={left},
legend entries={{\scriptsize vff},{\scriptsize sparse},{\scriptsize $M = 5$},{\scriptsize $M = 9$},{\scriptsize $M = 21$},{\scriptsize $M = 41$}},
legend style={draw=white!80.0!black}
]
\addlegendimage{only marks, mark=square*, color0}
\addlegendimage{only marks, mark=square*, color1}
\addlegendimage{only marks, mark=*, black}
\addlegendimage{only marks, mark=diamond*, black}
\addlegendimage{only marks, mark=triangle*, mark options={solid,rotate=180}, black}
\addlegendimage{only marks, mark=pentagon*, black}
\addlegendimage{only marks, mark=triangle*, black}
\addlegendimage{only marks, mark=square*, black}
\addlegendimage{only marks, mark=diamond*, mark options={solid,rotate=90}, black}
\addplot [semithick, color0, opacity=0.5, mark=*, mark size=3, mark options={solid}, only marks, forget plot]
table {0.126118183135986 16359.897873452
0.103002786636353 10263.7398867499
0.109531879425049 9653.81466854629
0.102914333343506 8524.80652681432
0.115167617797852 11667.7199737146
};
\addplot [semithick, color0, opacity=0.5, mark=diamond*, mark size=3, mark options={solid}, only marks, forget plot]
table {0.119397401809692 2871.25872529746
0.11044716835022 3204.14906717836
0.110409736633301 3549.03672100208
0.102834701538086 2890.88142572637
0.117263555526733 4102.4622855071
};
\addplot [semithick, color0, opacity=0.5, mark=triangle*, mark size=3, mark options={solid,rotate=180}, only marks, forget plot]
table {0.109511613845825 757.936938775478
0.128212451934814 945.332618277282
0.0960307121276856 742.469680449571
0.121248960494995 715.411595105368
0.109043836593628 725.667036436215
};
\addplot [semithick, color0, opacity=0.5, mark=pentagon*, mark size=3, mark options={solid}, only marks, forget plot]
table {0.130510091781616 473.682175338779
0.0956647396087646 461.120225043598
0.105266809463501 462.619870011661
0.0968248844146728 479.191252585973
0.0977797508239746 482.059170964742
};
\addplot [semithick, color1, opacity=0.5, mark=*, mark size=3, mark options={solid}, only marks, forget plot]
table {0.1408531665802 18400.1462798449
0.141074657440186 8977.05800387479
0.145119428634644 9281.97413664204
0.137389421463013 7671.03417215406
0.138490200042725 9567.7180056907
};
\addplot [semithick, color1, opacity=0.5, mark=diamond*, mark size=3, mark options={solid}, only marks, forget plot]
table {0.140424013137817 1980.81796691315
0.138952255249023 1791.44507995738
0.142049074172974 2513.31420676288
0.138784408569336 1698.6137364106
0.139683723449707 1834.98616421512
};
\addplot [semithick, color1, opacity=0.5, mark=triangle*, mark size=3, mark options={solid,rotate=180}, only marks, forget plot]
table {0.139354228973389 95.9336654144872
0.141758441925049 150.825654406298
0.140851497650146 138.245373113942
0.14207935333252 124.794706947338
0.141958713531494 135.723010963652
};
\addplot [semithick, color1, opacity=0.5, mark=pentagon*, mark size=3, mark options={solid}, only marks, forget plot]
table {0.145933628082275 13.7080406919349
0.144659280776978 13.912974658082
0.149548768997192 13.4185748469627
0.144931793212891 14.6179369945385
0.145194053649902 11.460862445238
};
\addplot [semithick, color0, mark=square*, mark size=3, mark options={solid}, only marks]
table {inf inf
};
\addplot [semithick, color1, mark=square*, mark size=3, mark options={solid}, only marks]
table {inf inf
};
\addplot [semithick, black, mark=*, mark size=3, mark options={solid}, only marks]
table {inf inf
};
\addplot [semithick, black, mark=diamond*, mark size=3, mark options={solid}, only marks]
table {inf inf
};
\addplot [semithick, black, mark=triangle*, mark size=3, mark options={solid,rotate=180}, only marks]
table {inf inf
};
\addplot [semithick, black, mark=pentagon*, mark size=3, mark options={solid}, only marks]
table {inf inf
};
\end{axis}

\end{tikzpicture}
     \caption{Input dimension $d = 1$}
  \end{subfigure}    \begin{subfigure}[t]{0.49\textwidth}
        \begin{tikzpicture}

\definecolor{color0}{rgb}{0.12156862745098,0.466666666666667,0.705882352941177}
\definecolor{color1}{rgb}{1,0.498039215686275,0.0549019607843137}

\begin{axis}[
xlabel={time (s)},
xmin=0.166135005206076, xmax=6.00738007336352,
ymin=-2986.59211098092, ymax=63096.3121699015,
xmode=log,
width=\figurewidth,
height=\figureheight,
tick align=outside,
tick pos=left,
x grid style={white!69.019607843137251!black},
y grid style={white!69.019607843137251!black},
legend entries={{\scriptsize vff},{\scriptsize sparse},{\scriptsize $M = 3^2$},{\scriptsize $M = 7^2$},{\scriptsize $M = 15^2$},{\scriptsize $M = 21^2$},{\scriptsize $M = 27^2$},{\scriptsize $M = 35^2$},{\scriptsize $M = 45^2$}},
legend cell align={left},
legend style={draw=white!80.0!black}
]
\addlegendimage{only marks, mark=square*, color0}
\addlegendimage{only marks, mark=square*, color1}
\addlegendimage{only marks, mark=*, black}
\addlegendimage{only marks, mark=diamond*, black}
\addlegendimage{only marks, mark=triangle*, mark options={solid,rotate=180}, black}
\addlegendimage{only marks, mark=pentagon*, black}
\addlegendimage{only marks, mark=triangle*, black}
\addlegendimage{only marks, mark=square*, black}
\addlegendimage{only marks, mark=diamond*, mark options={solid,rotate=90}, black}
\addplot [semithick, color0, opacity=0.5, mark=*, mark size=3, mark options={solid}, only marks, forget plot]
table {0.202995538711548 40126.5788570128
0.222471714019775 53528.0055632667
0.221302032470703 49950.649614579
0.220615863800049 44531.4857441453
0.195564270019531 50861.5304862881
};
\addplot [semithick, color0, opacity=0.5, mark=diamond*, mark size=3, mark options={solid}, only marks, forget plot]
table {0.216501474380493 10750.7459726247
0.215298652648926 10910.7151489109
0.200514554977417 10173.5581322939
0.209829807281494 9896.77126458017
0.210755586624146 10785.5971431891
};
\addplot [semithick, color0, opacity=0.5, mark=triangle*, mark size=3, mark options={solid,rotate=180}, only marks, forget plot]
table {0.216950416564941 2472.2014952326
0.2129967212677 2271.13198620493
0.223787546157837 2441.58720465986
0.217287302017212 2232.36060429155
0.198013067245483 2492.0104109755
};
\addplot [semithick, color0, opacity=0.5, mark=pentagon*, mark size=3, mark options={solid}, only marks, forget plot]
table {0.230341672897339 1473.74042487089
0.232399702072144 1451.45026290509
0.230613946914673 1461.9706096967
0.224767446517944 1381.95928894597
0.221921920776367 1431.68849043758
};
\addplot [semithick, color0, opacity=0.5, mark=triangle*, mark size=3, mark options={solid}, only marks, forget plot]
table {0.263471126556396 1183.54079193852
0.272954940795898 1126.82918254673
0.271883487701416 1126.63328209182
0.278165102005005 1108.5033193118
0.279739379882812 1125.98362287171
};
\addplot [semithick, color0, opacity=0.5, mark=square*, mark size=3, mark options={solid}, only marks, forget plot]
table {0.361517906188965 985.872348166478
0.359622478485107 966.366349209729
0.363096714019775 990.394216765771
0.364807367324829 963.958096898276
0.367379426956177 990.06410383471
};
\addplot [semithick, color0, opacity=0.5, mark=diamond*, mark size=3, mark options={solid, rotate=90}, only marks, forget plot]
table {0.68710732460022 920.767420786811
0.692674160003662 911.927266446215
0.684285163879395 933.468870979463
0.671258211135864 908.858547228663
0.677809000015259 924.157327902132
};
\addplot [semithick, color1, opacity=0.5, mark=*, mark size=3, mark options={solid}, only marks, forget plot]
table {0.199893951416016 43835.2139665018
0.201001405715942 60092.5437934977
0.199915647506714 55441.9959888867
0.200108528137207 49645.1315176445
0.200709819793701 56400.2942463051
};
\addplot [semithick, color1, opacity=0.5, mark=diamond*, mark size=3, mark options={solid}, only marks, forget plot]
table {0.212488174438477 7887.44691797624
0.215525150299072 8365.31756746154
0.21683406829834 7504.77430050099
0.212605714797974 7967.6400137227
0.210556745529175 7837.45574578815
};
\addplot [semithick, color1, opacity=0.5, mark=triangle*, mark size=3, mark options={solid,rotate=180}, only marks, forget plot]
table {0.305462121963501 751.590678550885
0.307155847549438 735.298100705582
0.304883480072021 724.136530369626
0.309434413909912 648.909142660536
0.313000679016113 739.965618005299
};
\addplot [semithick, color1, opacity=0.5, mark=pentagon*, mark size=3, mark options={solid}, only marks, forget plot]
table {0.498811483383179 258.090347478465
0.499816179275513 221.98411803041
0.499226808547974 224.919543263369
0.504605770111084 216.226951449897
0.503960609436035 216.713826453059
};
\addplot [semithick, color1, opacity=0.5, mark=triangle*, mark size=3, mark options={solid}, only marks, forget plot]
table {0.917831420898438 99.3852534317052
0.919410467147827 92.643513544881
0.919479370117188 95.5090591857852
0.919205904006958 97.7509862909774
0.921430587768555 99.8099546483868
};
\addplot [semithick, color1, opacity=0.5, mark=square*, mark size=3, mark options={solid}, only marks, forget plot]
table {2.03677701950073 45.7842092738429
2.02052354812622 42.6622809172732
2.03792238235474 44.3763642407762
2.0226104259491 44.9279035167328
2.01542735099793 42.5885430385101
};
\addplot [semithick, color1, opacity=0.5, mark=diamond*, mark size=3, mark options={solid, rotate=90}, only marks, forget plot]
table {5.04308176040649 19.1612372016389
5.08640217781067 18.6333632536366
5.05486392974854 18.6934376959298
5.08914303779602 18.79778083987
5.10336637496948 17.1762654228232
};
\addplot [semithick, color0, mark=square*, mark size=3, mark options={solid}, only marks]
table {inf inf
};
\addplot [semithick, color1, mark=square*, mark size=3, mark options={solid}, only marks]
table {inf inf
};
\addplot [semithick, black, mark=*, mark size=3, mark options={solid}, only marks]
table {inf inf
};
\addplot [semithick, black, mark=diamond*, mark size=3, mark options={solid}, only marks]
table {inf inf
};
\addplot [semithick, black, mark=triangle*, mark size=3, mark options={solid,rotate=180}, only marks]
table {inf inf
};
\addplot [semithick, black, mark=pentagon*, mark size=3, mark options={solid}, only marks]
table {inf inf
};
\addplot [semithick, black, mark=triangle*, mark size=3, mark options={solid}, only marks]
table {inf inf
};
\addplot [semithick, black, mark=square*, mark size=3, mark options={solid}, only marks]
table {inf inf
};
\addplot [semithick, black, mark=diamond*, mark size=3, mark options={solid}, only marks]
table {inf inf
};
\end{axis}

\end{tikzpicture}
     \caption{Input dimension $d = 2$}
  \end{subfigure} \\
    \begin{subfigure}[t]{0.49\textwidth}
        \begin{tikzpicture}

\definecolor{color0}{rgb}{0.12156862745098,0.466666666666667,0.705882352941177}
\definecolor{color1}{rgb}{1,0.498039215686275,0.0549019607843137}

\begin{axis}[
xlabel={time (s)},
ylabel={$\textsc{KL}\big[q\big(f(x)\big) || p\big(f(x)\given \by\big)\big]$},
xmin=0.22858046143266, xmax=7.44625908625624,
ymin=-2064.56889786084, ymax=75437.6811308141,
xmode=log,
width=\figurewidth,
height=\figureheight,
tick align=outside,
tick pos=left,
x grid style={white!69.019607843137251!black},
y grid style={white!69.019607843137251!black},
legend cell align={left},
legend style={draw=white!80.0!black},
legend entries={{\scriptsize vff},{\scriptsize sparse},{\scriptsize $M = 3^3$},{\scriptsize $M = 5^3$},{\scriptsize $M = 7^3$},{\scriptsize $M = 9^3$},{\scriptsize $M = 11^3$},{\scriptsize $M = 13^3$}}
]
\addlegendimage{only marks, mark=square*, color0}
\addlegendimage{only marks, mark=square*, color1}
\addlegendimage{only marks, mark=*, black}
\addlegendimage{only marks, mark=diamond*, black}
\addlegendimage{only marks, mark=triangle*, mark options={solid,rotate=180}, black}
\addlegendimage{only marks, mark=pentagon*, black}
\addlegendimage{only marks, mark=triangle*, black}
\addlegendimage{only marks, mark=square*, black}
\addlegendimage{only marks, mark=diamond*, mark options={solid,rotate=90}, black}
\addplot [semithick, color0, opacity=0.5, mark=*, mark size=3, mark options={solid}, only marks, forget plot]
table {0.284640312194824 56742.4415556632
0.275135278701782 63188.3191102639
0.294597387313843 61645.5514169603
0.285591840744018 55242.7972879885
0.287074565887451 58966.8640819829
};
\addplot [semithick, color0, opacity=0.5, mark=diamond*, mark size=3, mark options={solid}, only marks, forget plot]
table {0.291885375976562 28561.6427423447
0.294873476028442 28645.4475134608
0.277961254119873 30058.6892110832
0.294010162353516 28190.7457579465
0.277422666549683 29286.3876808687
};
\addplot [semithick, color0, opacity=0.5, mark=triangle*, mark size=3, mark options={solid,rotate=180}, only marks, forget plot]
table {0.306817054748535 14603.9699007781
0.300836801528931 15220.9068382087
0.288926362991333 15466.6610621525
0.299250602722168 14243.5754528311
0.29680609703064 15225.8031887936
};
\addplot [semithick, color0, opacity=0.5, mark=pentagon*, mark size=3, mark options={solid}, only marks, forget plot]
table {0.353863716125488 8779.23398352044
0.347698450088501 8862.12786425395
0.349225044250488 9195.8442455747
0.345361948013306 8796.73755441365
0.358263254165649 9209.88665554214
};
\addplot [semithick, color0, opacity=0.5, mark=triangle*, mark size=3, mark options={solid}, only marks, forget plot]
table {0.474977970123291 5798.85815655032
0.476574182510376 5906.40567950929
0.478173494338989 6075.41855385003
0.474207162857056 6089.32511576375
0.488597393035889 6155.42330314575
};
\addplot [semithick, color0, opacity=0.5, mark=square*, mark size=3, mark options={solid}, only marks, forget plot]
table {0.878152132034302 4206.86937784136
0.880542516708374 4286.45610031922
0.866629838943481 4339.73862669825
0.863762617111206 4427.87179068017
0.884839057922363 4337.76901033415
};
\addplot [semithick, color1, opacity=0.5, mark=*, mark size=3, mark options={solid}, only marks, forget plot]
table {0.268639087677002 64245.8180895331
0.268280982971191 66085.409236994
0.267797946929932 69531.3500744057
0.269484758377075 61181.462137229
0.30809473991394 71914.8515840562
};
\addplot [semithick, color1, opacity=0.5, mark=diamond*, mark size=3, mark options={solid}, only marks, forget plot]
table {0.323663949966431 25756.0162290517
0.319835424423218 27220.249422216
0.310817718505859 27472.6798448429
0.312576293945312 26184.8507332281
0.314158439636231 27346.7171040044
};
\addplot [semithick, color1, opacity=0.5, mark=triangle*, mark size=3, mark options={solid,rotate=180}, only marks, forget plot]
table {0.498146533966065 10341.0519447877
0.499467134475708 11319.2219925857
0.492123365402222 11461.0826766784
0.49664831161499 10212.0115666244
0.496087789535523 10766.4785387333
};
\addplot [semithick, color1, opacity=0.5, mark=pentagon*, mark size=3, mark options={solid}, only marks, forget plot]
table {1.05616617202759 4808.27552003829
1.0419487953186 4714.73255310804
1.05722117424011 5190.37474370881
1.04935216903687 5264.61314325585
1.04540491104126 5316.88890896661
};
\addplot [semithick, color1, opacity=0.5, mark=triangle*, mark size=3, mark options={solid}, only marks, forget plot]
table {2.51785612106323 2602.66858345924
2.54249238967896 2549.28079457925
2.51004552841186 2602.79932739997
2.57931208610535 2570.22964061884
2.5379204750061 2639.90911869989
};
\addplot [semithick, color1, opacity=0.5, mark=square*, mark size=3, mark options={solid}, only marks, forget plot]
table {6.25688290596008 1458.26064889712
6.27526068687439 1506.38016068193
6.25379109382629 1526.74710575031
6.19404029846191 1466.51726667706
6.35579681396484 1613.09003794827
};
\addplot [semithick, color0, mark=square*, mark size=3, mark options={solid}, only marks]
table {inf inf
};
\addplot [semithick, color1, mark=square*, mark size=3, mark options={solid}, only marks]
table {inf inf
};
\addplot [semithick, black, mark=*, mark size=3, mark options={solid}, only marks]
table {inf inf
};
\addplot [semithick, black, mark=diamond*, mark size=3, mark options={solid}, only marks]
table {inf inf
};
\addplot [semithick, black, mark=triangle*, mark size=3, mark options={solid,rotate=180}, only marks]
table {inf inf
};
\addplot [semithick, black, mark=pentagon*, mark size=3, mark options={solid}, only marks]
table {inf inf
};
\addplot [semithick, black, mark=triangle*, mark size=3, mark options={solid}, only marks]
table {inf inf
};
\addplot [semithick, black, mark=square*, mark size=3, mark options={solid}, only marks]
table {inf inf
};
\end{axis}

\end{tikzpicture}
     \caption{Input dimension $d = 3$}
  \end{subfigure}    \begin{subfigure}[t]{0.49\textwidth}
    \begin{tikzpicture}

\definecolor{color0}{rgb}{0.12156862745098,0.466666666666667,0.705882352941177}
\definecolor{color1}{rgb}{1,0.498039215686275,0.0549019607843137}

\begin{axis}[
xlabel={time (s)},
xmin=0.276907042645998, xmax=77.6299619758172,
ymin=2132.59678583347, ymax=81233.6784053978,
xmode=log,
width=\figurewidth,
height=\figureheight,
tick align=outside,
tick pos=left,
x grid style={white!69.019607843137251!black},
y grid style={white!69.019607843137251!black},
legend cell align={left},
legend style={draw=white!80.0!black},
legend entries={{\scriptsize vff},{\scriptsize sparse},{\scriptsize $M = 3^4$},{\scriptsize $M = 5^4$},{\scriptsize $M = 7^4$},{\scriptsize $M = 9^4$}}
]
\addlegendimage{only marks, mark=square*, color0}
\addlegendimage{only marks, mark=square*, color1}
\addlegendimage{only marks, mark=*, black}
\addlegendimage{only marks, mark=diamond*, black}
\addlegendimage{only marks, mark=triangle*, mark options={solid,rotate=180}, black}
\addlegendimage{only marks, mark=pentagon*, black}
\addlegendimage{only marks, mark=triangle*, black}
\addlegendimage{only marks, mark=square*, black}
\addlegendimage{only marks, mark=diamond*, mark options={solid,rotate=90}, black}
\addplot [semithick, color0, opacity=0.5, mark=*, mark size=3, mark options={solid}, only marks, forget plot]
table {0.362338066101074 66102.8267266571
0.378393888473511 69703.5859193085
0.364611864089966 62892.2410205047
0.369531631469727 67203.6724786194
0.363287687301636 70767.6663646623
};
\addplot [semithick, color0, opacity=0.5, mark=diamond*, mark size=3, mark options={solid}, only marks, forget plot]
table {0.413476705551147 35765.3095810963
0.415077924728394 36524.7528295168
0.418456554412842 35671.2174161281
0.412836790084839 35035.0362058217
0.415892124176025 36604.3413431653
};
\addplot [semithick, color0, opacity=0.5, mark=triangle*, mark size=3, mark options={solid,rotate=180}, only marks, forget plot]
table {1.12315464019775 18760.1763194757
1.10734248161316 18818.7465504687
1.09573674201965 18830.9898582351
1.08369636535645 18654.4441831365
1.08307003974915 19029.9995980855
};
\addplot [semithick, color0, opacity=0.5, mark=pentagon*, mark size=3, mark options={solid}, only marks, forget plot]
table {9.41337823867798 11285.0568929649
9.35542917251587 11225.7002854531
9.39137601852417 11072.0924832038
9.32186794281006 10901.7000160729
9.35824513435364 11218.2451630952
};
\addplot [semithick, color1, opacity=0.5, mark=*, mark size=3, mark options={solid}, only marks, forget plot]
table {0.358903408050537 75590.1018263915
0.366266489028931 77638.1746954176
0.357760906219482 70568.0783276859
0.359489917755127 74866.7735450161
0.366631507873535 75756.8559718154
};
\addplot [semithick, color1, opacity=0.5, mark=diamond*, mark size=3, mark options={solid}, only marks, forget plot]
table {1.00599265098572 31586.747603087
0.988969087600708 32686.9465158816
0.990988969802856 32298.0958690584
1.0125560760498 32096.3410148572
0.999912261962891 33108.3421008972
};
\addplot [semithick, color1, opacity=0.5, mark=triangle*, mark size=3, mark options={solid,rotate=180}, only marks, forget plot]
table {7.74358606338501 13070.044949433
7.73823809623718 13187.3584096535
7.79894614219666 13029.2551107132
7.72540855407715 12590.9234491325
7.85310983657837 13180.1287789813
};
\addplot [semithick, color1, opacity=0.5, mark=pentagon*, mark size=3, mark options={solid}, only marks, forget plot]
table {59.3212592601776 6044.37432506319
58.903687953949 6068.39421510988
59.058260679245 5989.18753739419
58.9549443721771 5728.10049581367
60.0856125354767 5960.38559905695
};
\addplot [semithick, color0, mark=square*, mark size=3, mark options={solid}, only marks]
table {inf inf
};
\addplot [semithick, color1, mark=square*, mark size=3, mark options={solid}, only marks]
table {inf inf
};
\addplot [semithick, black, mark=*, mark size=3, mark options={solid}, only marks]
table {inf inf
};
\addplot [semithick, black, mark=diamond*, mark size=3, mark options={solid}, only marks]
table {inf inf
};
\addplot [semithick, black, mark=triangle*, mark size=3, mark options={solid,rotate=180}, only marks]
table {inf inf
};
\addplot [semithick, black, mark=pentagon*, mark size=3, mark options={solid}, only marks]
table {inf inf
};
\end{axis}

\end{tikzpicture}
     \caption{Input dimension $d = 4$}
  \end{subfigure} \\
    \caption{\label{fig:increase_dim}    KL divergence versus wall clock time for sparse GPs and VFF with various number $M$ of inducing inputs. The data is obtained by sampling a GP with \maternthreetwo kernel ($\sigma^2=1,\ \ell=0.2$) at $10^4$ uniformly distributed input locations and $\cN(0,0.1)$ observation noise. The KL is computed with respect to the full GP regression model with known parameters for all models. The experiments are replicated five times with different draws of $\bX$ and $\by$.}
\end{figure}
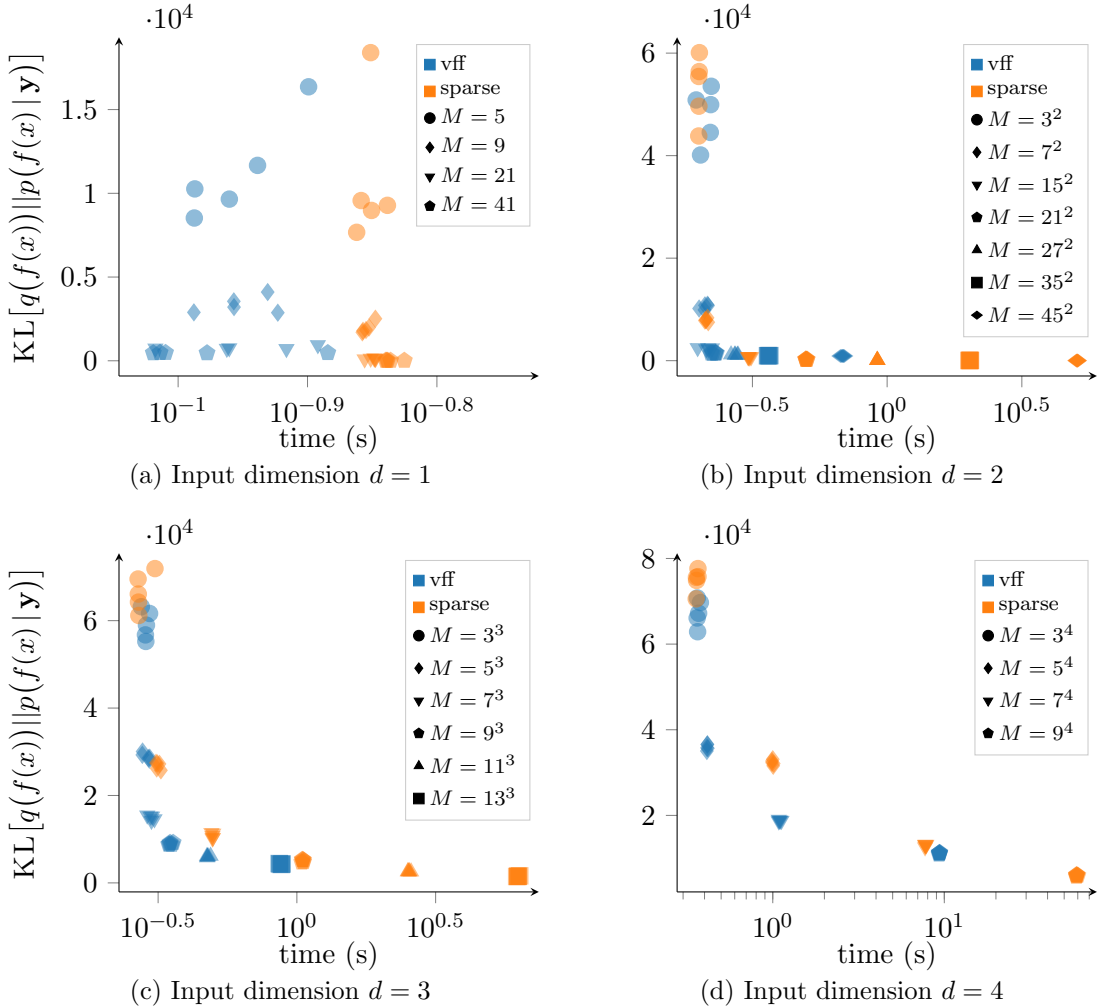

The results are reported in Figure~\ref{fig:increase_dim}. In one dimension, both methods have a similar accuracy in term of KL for a given number of inducing variables but VFF appears to be faster than sparse GPs. However the low complexity of this one dimensional model does not allow us to measure the time accurately enough to distinguish the influence of the number of inducing variables on the execution time. More interesting conclusions can be drawn when the input space is greater than one: for the same number of inducing variables, VFF is significantly faster than the sparse GP (one fewer Cholesky decomposition is required) but the quality of the approximation is not as good as the sparse GP for the same number of basis functions.

For a given computational time, both methods have an equivalent accuracy but with a different number of inducing variables. The VFF basis functions appear to have less representational power, but their structure allows faster computations. Our understanding is that this smaller representational power of VFF can be explained by the fact that the inducing functions need to account for variations on $[a,b]^d$ whereas the sparse GP basis functions only need account for that on $[0,1]^d$: a much smaller volume, especially in high dimensions. Note that the accuracy of the sparse GP can be improved by optimization of the location of the inducing inputs but that would result in computation times many times larger.

We have shown empirically that in the {\em worst case}, VFF is comparable to the sparse GP in terms of the available computate-time/accuracy tradeoff. For other models including additive models, we expect much stronger performance from the VFF approximation, as the following sections demonstrate.

\subsection{Additive modelling of airline delays}
We demonstrate the feasibility of the variational spectral approximation in a high-dimensional GP regression example for predicting airline delays. The US flight delay prediction example \citep[see][for the original example]{hensman2013gaussian} has reached the status of a standard test dataset in Gaussian process regression, partly because of its large-scale non-stationary nature and partly because of the massive size of the dataset with nearly 6~million records.

This example has recently been used by \citet{deisenroth2015distributed}, where it was solved using distributed Gaussian processes. \citet{samo2016string} use this example for demonstrating the computational efficiency of string Gaussian processes. Furthermore, \citet{adam2016scalable} bring this problem forward as an example of a dataset, where the model can be formed by the addition of multiple underlying components.

The dataset consists of flight arrival and departure times for every commercial flight in the USA for the year 2008. Each record is complemented with details on the flight and the aircraft.  The eight covariates $\bx$ are the same as in \citet{hensman2013gaussian}, namely the age of the aircraft (number of years since deployment), route distance, airtime, departure time, arrival time, day of the week, day of the month, and month. We predict the delay of the aircraft at landing (in minutes), $y$.

Following the proposed solution by \citet{adam2016scalable} for this estimation problem, we use a Gaussian process regression model with a prior covariance structure given as a sum of \maternthreetwo covariance functions for each input dimension, and assume the observations to be corrupted by independent Gaussian noise, $\varepsilon_i \sim \mathcal{N}(0,\sigma_n^2)$. The model is
\begin{align}
    f(\bx) &\sim \mathcal{GP}\left(0, \sum_{d=1}^8 k(x_d, x_d')\right)\,,\\
    y_i &= f(\bx_i) + \varepsilon_i\,,
\end{align}
for $i=1,2,\ldots,N$. For the Variational Fourier Feature method (VFF), we used $M=30$ frequencies per input dimension.

We consider the entire dataset of 5.93~million records. To provide comparable predictions, we follow the test setup in the paper by \citet{samo2016string}, where results were calculated using String GPs (their method), the Bayesian committee machine \citep[BCM,][]{tresp2000bayesian}, the robust Bayesian committee machine \citep[rBCM,][]{deisenroth2015distributed}, and stochastic variational GP inference \citep[SVIGP,][]{hensman2013gaussian}. The SVIGP used a squared-exponential kernel with one lengthscale per dimension. We repeated the SVIGP experiments of \citet{samo2016string} to complete the table for larger datasets and predictive densities.

Predictions are made for several subset sizes of the data, each selected uniformly at random: 10,000, 100,000, 1,000,000, and 5,929,413 (all data). In each case, two thirds of the data is used for training and one third for testing. For each subset size, the training is repeated ten times, and for each run, the outputs are normalized by subtracting the training sample mean from the outputs and dividing the result by the sample standard deviation. In the VFF method, the inputs were normalized to $[0, 1]$ and the domain size was set to $[-2, 3]$, i.e.\ the boundary is at a distance of two times the range of the data for each dimension.

\begin{table}
  \centering
  \resizebox{\textwidth}{!}{
  \begin{tabular}{lcccccccc}
\toprule
$N$ & \multicolumn{2}{c}{10,000} & \multicolumn{2}{c}{100,000} & \multicolumn{2}{c}{1,000,000} & \multicolumn{2}{c}{5,929,413} \\
& MSE & \add{NLPD} & MSE & \add{NLPD} & MSE & \add{NLPD} & MSE & \add{NLPD} \\
\midrule
VFF & 0.89 $\pm$ 0.15 & \add{1.362 $\pm$ 0.091} & 0.82 $\pm$ 0.05 & \add{1.319 $\pm$ 0.030} & 0.83 $\pm$ 0.01 & \add{1.326 $\pm$ 0.008} & 0.827 $\pm$ 0.004 & \add{1.324 $\pm$ 0.003} \\
Full-RBF & 0.89 $\pm$ 0.16 & \add{1.349 $\pm$ 0.098} & N/A & \add{N/A} & N/A & \add{N/A} & N/A & \add{N/A} \\
Full-additive & 0.89 $\pm$ 0.16 & \add{1.362 $\pm$ 0.096} & N/A & \add{N/A} & N/A & \add{N/A} & N/A & \add{N/A} \\
SVIGP & \add{0.89 $\pm$ 0.16} & \add{1.354 $\pm$ 0.096} & \add{0.79 $\pm$ 0.05} & \add{1.299 $\pm$ 0.033} & \add{0.79 $\pm$ 0.01} & \add{1.301 $\pm$ 0.009} & \add{0.791 $\pm$ 0.005} & \add{1.300 $\pm$ 0.003} \\
String GP$^\dagger$ & 1.03 $\pm$ 0.10 & \add{N/A} & 0.93 $\pm$ 0.03 & \add{N/A} & 0.93 $\pm$ 0.01 & \add{N/A} & 0.90 $\pm$ 0.01 & \add{N/A} \\
rBCM$^\dagger$ & 1.06 $\pm$ 0.10 & \add{N/A} & 1.04 $\pm$ 0.04 & \add{N/A} & N/A & \add{N/A} & N/A & \add{N/A} \\
\bottomrule
  \end{tabular}  }
  \caption{\label{tbl:airline}
  Predictive mean squared errors (MSEs) and \add{negative log predictive densities (NLPDs) with} one standard deviation on the airline arrival delays experiment.  $^\dagger$MSE results as reported by \citet{samo2016string}, who also list results for the BCM and rBCM methods.}
\end{table}

Table~\ref{tbl:airline} shows the predictive mean squared errors (MSEs) and \add{the negative log predictive densities (NLPDs) with} one standard deviation on the airline arrival delays experiment. In this form the MSEs over the normalized data can be interpreted as a fraction of the sample variance of airline arrival delays. Thus a MSE of $1.00$ is \remove{as good as} \add{an accuracy equivalent to} using the training mean as predictor. The results for the String GP method are included for reference only and they are given as reported by \citet{samo2016string}. They also list results for the rBCM method, which performed worse than the String GP.

Running the VFF experiment with all $5.93$~million data using our Python implementation took $265 \pm 6$~seconds ($626\pm11$~s CPU time) on a two-core MacBook Pro laptop (with all calculation done on the CPU). The results for 10,000 data were calculated in $21\pm2$~seconds ($27\pm4$~s CPU time). For comparison, according to \citet{samo2016string}, running the String GP experiment took 91.0~hours total CPU time (or 15~h of wall-clock time on an 8-core machine). \add{Even the SVIGP method required $5.1\pm0.1$~hours of computing ($27.0\pm0.8$~h CPU time) on a cluster.}

For comparison, we have also included the results for the naive full GP solution for 10,000 data. These results were computed on a computing cluster and the computation time was several hours per repetition. We show results both for a squared-exponential kernel with one lengthscale per dimension, and also with an additive kernel similar to the one defined for the VFF method. As one might expect, the RBF model had a slightly lower average MSE ($0.8898$) than the additive model ($0.89274$). The difference is small: only 3~seconds of RMSE in the original units of flight-delay. The VFF result ($0.8934$) for the same data is close to the full GP solution (the difference being only about 0.66~s in terms of non-normalized RMSE). \add{In terms of NLPD the results agree as well; the full RBF model is able to capture the phenomenon slightly better than the full additive model, and the VFF model approximates well the full additive model.} The variability in the data is large, and including more data in the GP training gives the model more power for prediction. \add{The SVIGP model (with a squared-exponential kernel) is the most flexible one and reaches the lowest RMSE and NLPD in the experiments.}

Figure~\ref{fig:airline} shows the component-wise predictions for each of the covariates, thus summarizing the underlying structure of the delay prediction model. As one might expect, the effect of the month, day of month, and day of week are small. The only interpretable effects are seen in slightly higher delays in the summer months and around the holiday season towards the end of the year, as well as some effects for the weekend traffic on Friday and Sunday. The age of the aircraft has a barely noticeable effect. The four remaining inputs (e--h) explain most of the delay. The effects of airtime and travel distance are  strongly correlated. Strong correlation can also be seen in departure and arrival time effects, both of which show peaks during the night. The model also catches the periodic nature of the departure and arrival time effects. These results are in line with the analysis of \citet{adam2016scalable}.

\newcommand\mysubfigwidth{0.48\textwidth}
\begin{figure}
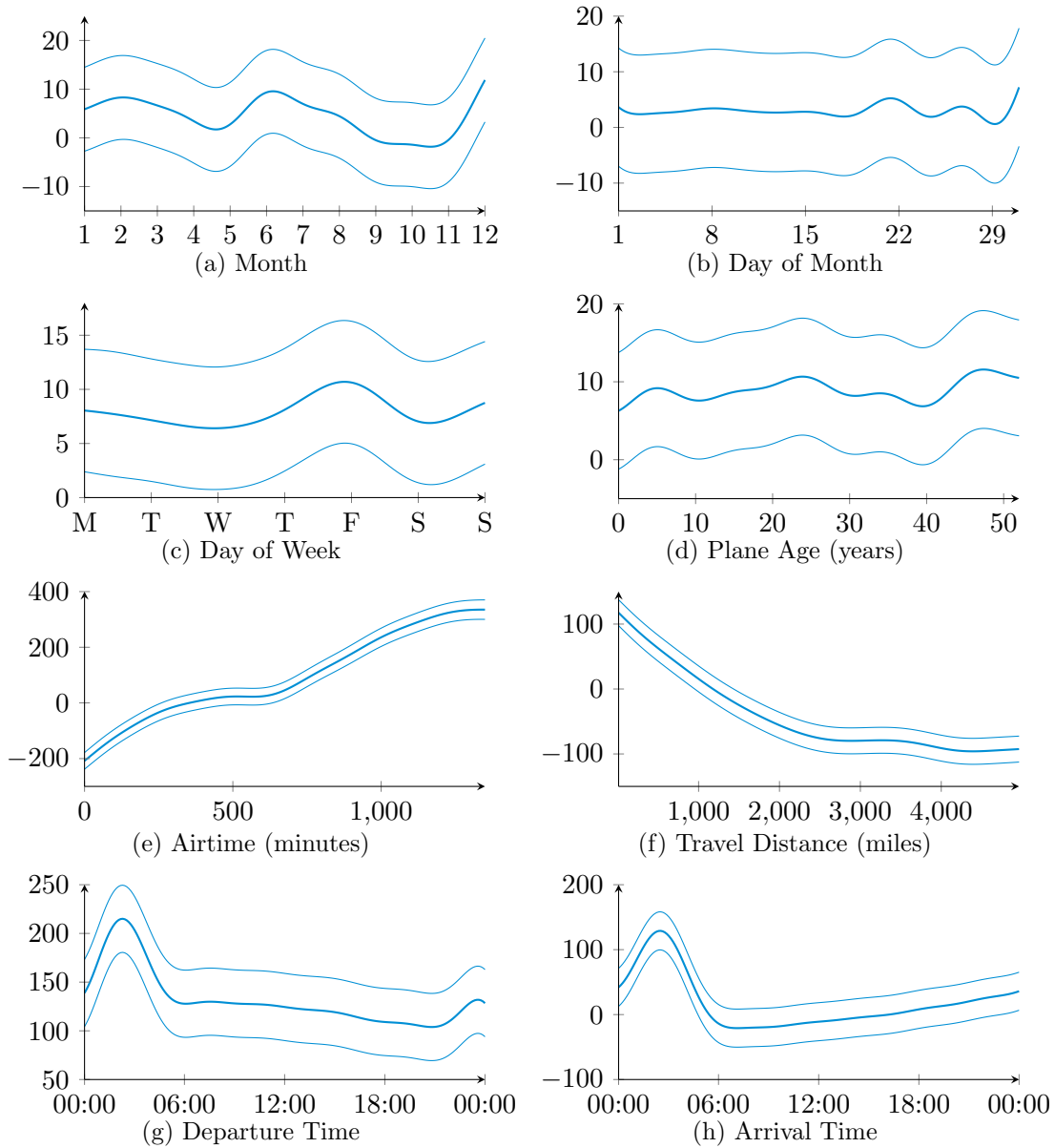

  \captionsetup[subfigure]{aboveskip=-0.3ex,belowskip=0.3ex}
  \centering
  \setlength\figurewidth{0.47\textwidth}
  \setlength\figureheight{0.6\figurewidth}
      \pgfplotsset{yticklabel style={text width=1cm, align=right}}
    \begin{subfigure}[t]{\mysubfigwidth}
    \pgfplotsset{xtick={1,2,3,4,5,6,7,8,9,10,11,12}}

     \caption{Month}
  \end{subfigure}    \begin{subfigure}[t]{\mysubfigwidth}
    \pgfplotsset{xtick={1,8,15,22,29}}
    %
     \caption{Day of Month}
  \end{subfigure} \\
    \begin{subfigure}[t]{\mysubfigwidth}
    \pgfplotsset{xtick={1,2,3,4,5,6,7},xticklabels={M,T,W,T,F,S,S}}
    %
     \caption{Day of Week}
  \end{subfigure}    \begin{subfigure}[t]{\mysubfigwidth}
    %
     \caption{Plane Age (years)}
  \end{subfigure} \\
    \begin{subfigure}[t]{\mysubfigwidth}
    \pgfplotsset{xtick={0, 500, 1000, 1500}}
    %
     \caption{Airtime (minutes)}
  \end{subfigure}    \begin{subfigure}[t]{\mysubfigwidth}
    %
     \caption{Travel Distance (miles)}
  \end{subfigure} \\
    \begin{subfigure}[t]{\mysubfigwidth}
    \pgfplotsset{xtick={1, 360, 720,  1080, 1440}, xticklabels={00:00,06:00,12:00,18:00,00:00}}
    %
     \caption{Departure Time}
  \end{subfigure}    \begin{subfigure}[t]{\mysubfigwidth}
    \pgfplotsset{xtick={1, 360, 720,  1080, 1440}, xticklabels={00:00,06:00,12:00,18:00,00:00}}
    %
     \caption{Arrival Time}
  \end{subfigure}
    \caption{\label{fig:airline}    The posterior mean and two posterior standard deviations in the airline delay prediction experiment estimated from 5.93~million data. Each panel show the posterior over the effect from one covariate under an additive model. The vertical axis represents delay in minutes in each case.}
\end{figure}

\subsection{Classification with a Gaussian approximation}
\label{par:experiment_classification}
\begin{figure}
    \setlength\figurewidth{0.35\textwidth}
    \setlength\figureheight{0.35\textwidth}
    \begin{center}
    \begin{tikzpicture}

\begin{groupplot}[
group style={group size=3 by 1,
             xticklabels at=edge bottom,
             yticklabels at=edge left}
]
\nextgroupplot[
axis y line*=left,
axis x line*=bottom,
xmin=0, xmax=25,
ymin=-315, ymax=-135,
axis on top,
width=\figurewidth,
height=\figureheight,
title={Full covariance},
xlabel={$M$},
ylabel={ELBO}
]
\addplot [only marks, blue538, mark=*, mark size=2]
table {2 -310.701953172
4 -151.529180963
6 -143.004411451
8 -140.609600376
10 -140.256859437
12 -140.033604939
14 -139.929551773
16 -140.225325452
};

\addplot [red538, thick]
table {0 -139.128
30 -139.128
};

\nextgroupplot[
axis y line*=left,
axis x line*=bottom,
xmin=0, xmax=25,
ymin=-315, ymax=-135,
axis on top,
width=\figurewidth,
height=\figureheight,
title={Kronecker},
xlabel={$M$}
]
\addplot [only marks, blue538, mark=*, mark size=2]
table {2 -3.111829875467602164e+02
4 -1.546238779293163077e+02
6 -1.487761106719686381e+02
8 -1.488393047745154263e+02
10 -1.507199931694538861e+02
12 -1.524523844360953433e+02
14 -1.540629741057444448e+02
16 -1.555421079740847858e+02
18 -1.568657641635780635e+02
20 -1.580469346906508576e+02
22 -1.591234442417904802e+02
24 -1.601019032982775911e+02
};
\addplot [red538, thick]
table {0 -139.128
30 -139.128
};

\nextgroupplot[
axis y line*=left,
axis x line*=bottom,
xmin=0, xmax=25,
ymin=-315, ymax=-135,
axis on top,
width=\figurewidth,
height=\figureheight,
title={Kronecker sum},
xlabel={$M$}
]
\addplot [only marks, blue538, mark=*, mark size=2]
table {2 -310.76471546
4 -152.10050351
6 -143.89958137
8 -141.63274004
10 -141.41810529
12 -141.15702598
14 -141.09586209
16 -141.0806787
18 -141.07271214
20 -141.04085493
22 -141.13322317
24 -141.13264877
};

\addplot [red538, thick]
table {0 -139.128
30 -139.128
};
\end{groupplot}

\end{tikzpicture}
     \end{center}
    \caption{\label{fig:banana_convergence}The increasing ELBO for the banana dataset with increasing numbers of frequencies $M$, and a Gaussian approximation to the posterior. Left: $q(\bu)$ has an unconstrained covariance matrix. Middle: $q(\bu)$ has a Kronecker-structured covariance matrix. Right: $q(\bu)$ has a covariance matrix with a sum of two Kronecker-structured matrices.  In each case the ELBO achieved using a full matrix decomposition is shown as a horizontal red line.}
\end{figure}
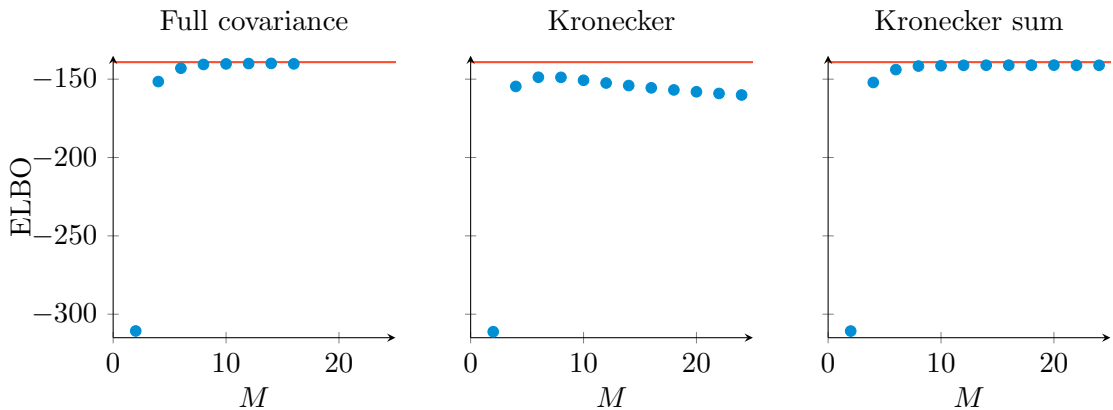
To perform variational classification using Variational Fourier Features, we can use a Gaussian approximation to the posterior $q(\bu)$ and optimize the ELBO with respect to the mean and variance of the approximation, as described in Section \ref{par:gaussian_approximations}.
To illustrate, we replicate the experiment on the banana dataset from \citet{hensman2015scalable}. In that work, the authors showed that a variational inducing point method performed better than competing alternatives because the positions of the inducing points could be effectively optimized using the ELBO as the objective. In this work, we have selected the frequencies of the Variational Fourier Features to be placed on a regular grid: the only aspect of the Fourier features to be tuned is $M$, controlling the number of features used.

Since the variational framework provides a guarantee that more inducing variables must be monotonically better \citep{titsias2009variational}, we expect that increasing the number of Variational Fourier Features will provide improved approximations. However, if we restrict the form of the Gaussian approximation by requiring a structured covariance, this may no longer be true. In practice, we have found that a freely-structured covariance matrix works well in one dimension and for additive models, but for Kronecker-structured models, optimization over an exponentially large matrix is intolerably slow. A suggestion for a related method provided by \citet{nickson2015blitzkriging} is to force the Gaussian approximation to have a Kronecker-structured covariance matrix. This reduces the number of variables over which we have to optimize, but removes the guarantee that more inducing variables is always better. We find in practice that this structure does not work well: Figure~\ref{fig:banana_convergence} shows the ELBO as a function of the number of inducing variables $M$, for both a trained covariance and a Kronecker--structured one. We see that for the Kronecker structure, the ELBO decreases as $M$ increases, implying that more inducing variables are actually making the approximation {\em worse} in this case.

Our suggestion is to use a sum of two Kronecker-structured matrices. The right-most plot of Figure~\ref{fig:banana_convergence} shows that the ELBO increases monotonically with $M$ when we use this structure. We provide no guarantee that the sum of two Kronecker matrices is optimal, and suggest that future work might consider generalizing the form of this approximation.

\begin{figure}
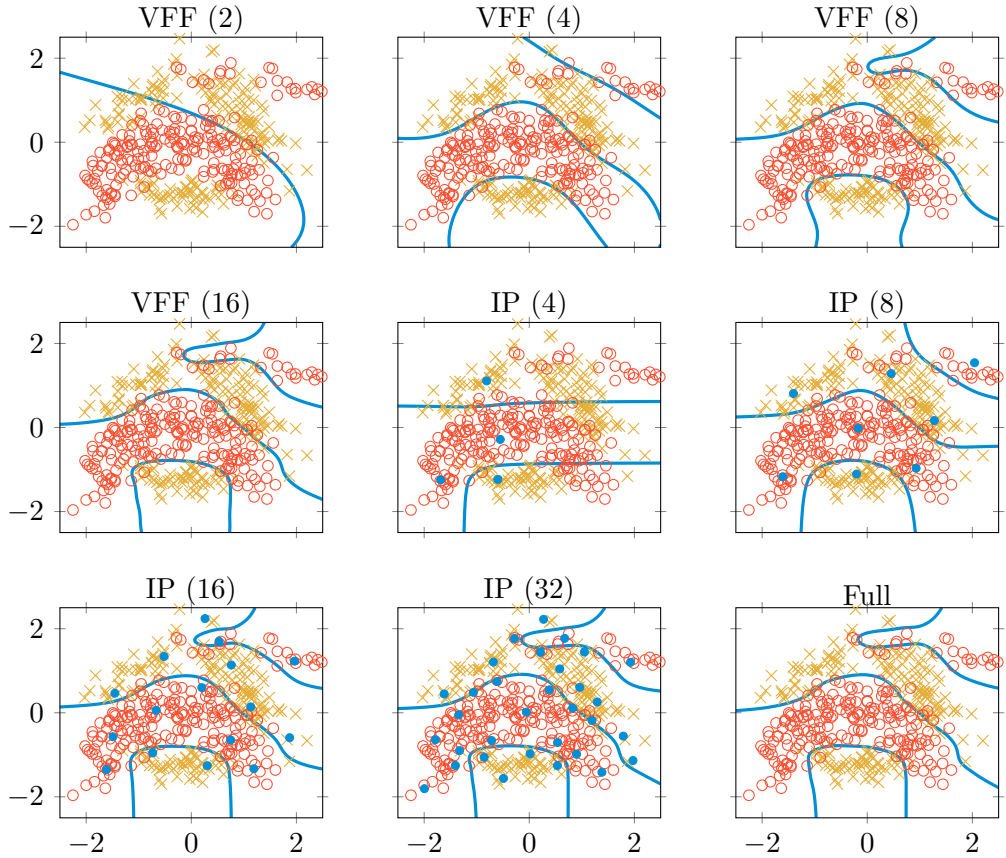

    \centering
 \caption{\label{fig:banana}
    Classification of the banana dataset with increasing number of Variational Fourier Features (VFF) and Inducing Points (IP). The two classes of data are shown as red circles and yellow crosses, and the decision boundary is shown as a solid blue line. An approximation with a full inversion of the covariance matrix is shown in the last plot. The blue dots show optimized positions of the inducing points.  The Variational Fourier Features approach the true posterior more rapidly than the inducing point method.
}
\end{figure}

To compare our method with the inducing-point method, we refer to Figure~\ref{fig:banana}, which
shows the results of fitting a Gaussian process classifier to the banana dataset using increasing numbers of inducing points (IP) and increasing numbers of Fourier features (VFF).  We observe the effect noted in \citet{hensman2015scalable} that the inducing points move toward the decision boundary, though some slight differences from that experiment exist because of our choice of the product of \maternfivetwo kernels, instead of the squared exponential kernel. The Figure also shows that the number of frequencies required to obtain a good approximation is reasonable, though it is not straightforward to compare directly between the VFF and IP methods. The total number of basis functions used in the Fourier features is twice the number of frequencies ($M$) listed, squared (since we have real and imaginary components, and take the Kronecker product). Still, the VFF method requires fewer computations per frequency than the IIP method does per inducing point, since the $\Kuu$ matrix is easily decomposed.

\subsection{Solar irradiance}
\begin{figure}
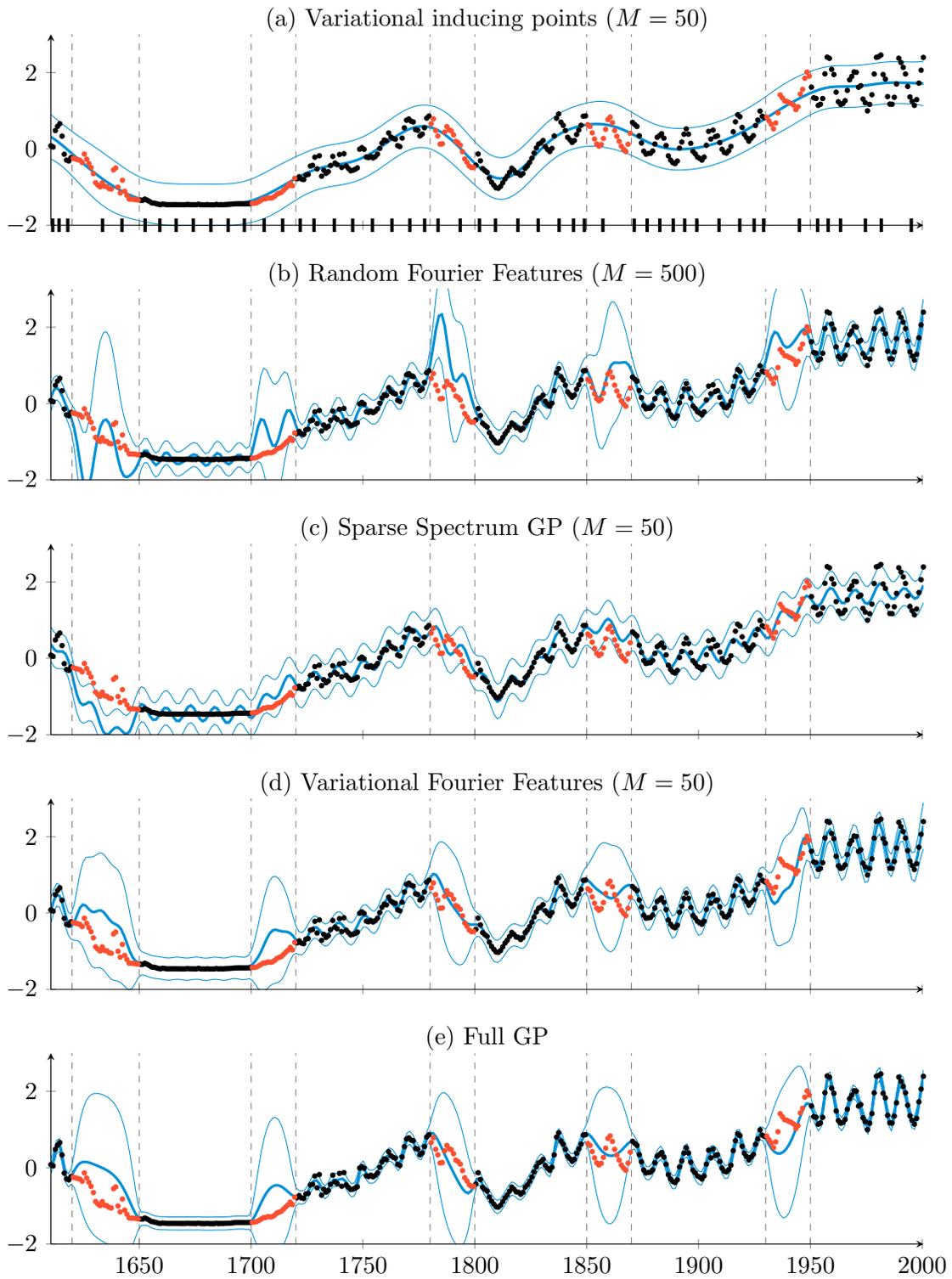

\centering
\pgfplotsset{x tick label style={/pgf/number format/.cd,fixed,precision=3, set thousands separator={}}}

 \caption{\label{fig:solar}
Replicating the experiment on solar irradiance data from \citet{gal2015improving}.}
\end{figure}

An alternative, but related idea to our proposed Variational Fourier Features approach is presented by \citet{gal2015improving}. In that work, the authors used a Random Fourier Features model (as in equation~\eqref{eq:rff_model}), but additionally performed a variational-Bayesian treatment of the frequencies. We repeat an experiment from that work on solar irradiance data \citep{lean2004}, shown in Figure~\ref{fig:solar}.
In this experiment, the time series was normalized and segments of the data were removed; in Figure~\ref{fig:solar}, removed data are shown in red, and vertical dashed lines denote the removed sections. The remaining data are shown in black, and we show the predicted mean (heavy blue line) and 95\% confidence intervals (light blue lines). We fitted five models (or model approximations) to the remaining data, using 50 inducing points (or frequencies) except in the case of the Random Fourier Features, where we used 500 \citep[following][]{gal2015improving}. In each case, we used a \maternfivetwo kernel initialized with a lengthscale of 10, and optimized the parameters with respect to the marginal likelihood (or \textsc{ELBO}, accordingly).

Figure~\ref{fig:solar}(a) shows the results of an inducing point method  \citep[still within the variational framework, based on][]{titsias2009variational}. The positions of the inducing points $z$ are shown as black marks on the horizontal axis. In this case, there are insufficient points to capture the complexity of the data, and the model reverts to a long lengthscale. This is an interesting case of underfitting, where the variational approximation to the function values gives rise to a bias in the estimate of the lengthscale parameter. This bias occurs because when maximising the \textsc{ELBO} with respect to the parameters, the \textsc{KL} divergence is smaller for longer lengthscales, biasing the parameter estimation in that direction. \citet{turner2011two} give an excellent discussion of this issue.

Figure~\ref{fig:solar}(b) shows the results of the Random Fourier Features method. This method is given 500 frequencies; we concur with the findings of \citet{gal2015improving} that using only 50 frequencies gives very poor performance. Where the data are missing, the error bars provided by the model are large, similarly to the full GP (Figure~\ref{fig:solar}(e)). The approximation is unable to recover the correct mean function, however, and is somewhat subject to overfitting: the model emphasises the high-frequency content in the data. In some sense, this is a desirable property as the model is more flexible than the original GP. For example, \citet{yang2015carte} exploited this property to fit complex GP models. However, we advocate the variational method precisely because of the asymptotic convergence (in $M$) to the true posterior distribution of the original model.

Figure~\ref{fig:solar}(c) shows the result of fitting the Sparse Spectrum GP approximation \citep{lazaro2010sparse}. This model is similar to the Random Fourier Features model, except that the frequencies are optimized. Here the overfitting is more severe: the model focusses entirely on the high-frequency content of the data. Although this gives good predictions in some cases (the segment around 1930 for example), the predictions are wild in others (e.g. around 1630, where none of the held-out data fall within the 95\% intervals).

Figure~\ref{fig:solar}(d) and (e) show the Variational Fourier Features method and the full GP method. The VFF approximation strives to approach the true posterior, and it is clear that the method is superior to the inducing points method in (a) in that the lengthscale is well estimated and the confidence intervals match the behaviour of the full GP. Some small discrepancies exist, but we note that these disappear completely at 100 inducing frequencies.

In summary, this experiment emphasises again how the variational approach attempts to approximate the posterior of the model, whereas previous approximations have {\it changed} the model, giving different behaviours. Within the variational methods, the Variational Fourier Features outperform the inducing point approach in this case, since our orthogonal basis functions are capable of capturing the posterior more effectively.

\subsection{MCMC approximations}
Above we have used the Gaussian approximation to the posterior process. \citet{hensman2015mcmc} propose a methodology which combines the variational approach with MCMC by sampling from the optimal approximating distribution, as we described in equation \eqref{eq:q_hat}. The advantages of this approach are that the posterior need not be assumed Gaussian, and that the covariance parameters can also be treated in a Bayesian fashion. The VFF framework that we have described fits well into this scheme, and as we have already described, the computational cost per iteration is linear in the number of data and in the number of required frequencies.

Here we replicate an experiment from \citet{hensman2015mcmc} (therein based on inducing point representations) in estimating the rate of a two-dimensional \remove{Log} \add{log} Gaussian Cox process. In this experiment, we aim to emphasise the convergent properties of Variational Fourier Features, and show how they might be used in practice within the MCMC framework.

The data consist of a series of 127 points collected in a square area, denoting the location of pine saplings. This simple illustration was also used by \citet[e.g.][]{moller1998log}. The model is
\begin{align}
    f(\bs) &\sim \mathcal{GP}\big(0, k(\bs, \bs')\big)\,,\\
    \log \lambda(\bs) &= f(\bs) + c\,,\\
    \by &\sim \mathcal {PP}(\lambda)\,,
\end{align}
where $\mathcal{PP}$ denotes an inhomogeneous Poisson process with intensity $\lambda$, $\by$ are spatial locations representing a draw from that Poisson process, $k$ is a product of \maternthreetwo covariance functions of the two-dimensional space indexed by $\bs$, and $c$ is a parameter to be inferred alongside the covariance function parameters. The model contains a doubly-intractable likelihood, which can be approximated by gridding the space and assuming that the intensity is constant across the bin size. The resulting approximation to the likelihood is
\begin{align}
    p(\by \given \bff) = \prod_{i=1}^{G}\mathcal{P}(N_i\given\lambda(\mathbf s_i) \Delta)\,,
\end{align}
where $G$ is the number of grid cells, $\mathbf s_i$ represent the \remove{centers} \add{centres} of those cells, $N_i$ is the number of data in the $i^\textrm{th}$ cell, $\Delta$ is the area of a grid cell and $\mathcal P$ represents the Poisson distribution. To perform inference in this model, we decompose the GP with Variational Fourier features, use the rotation described in Section~\ref{par:rot_mcmc} to \add{centre} the $\bu$ variables and run Hamiltonian Monte Carlo (HMC) to jointly sample the \add{centred} variables alongside the covariance function parameters and $c$.

\begin{figure}
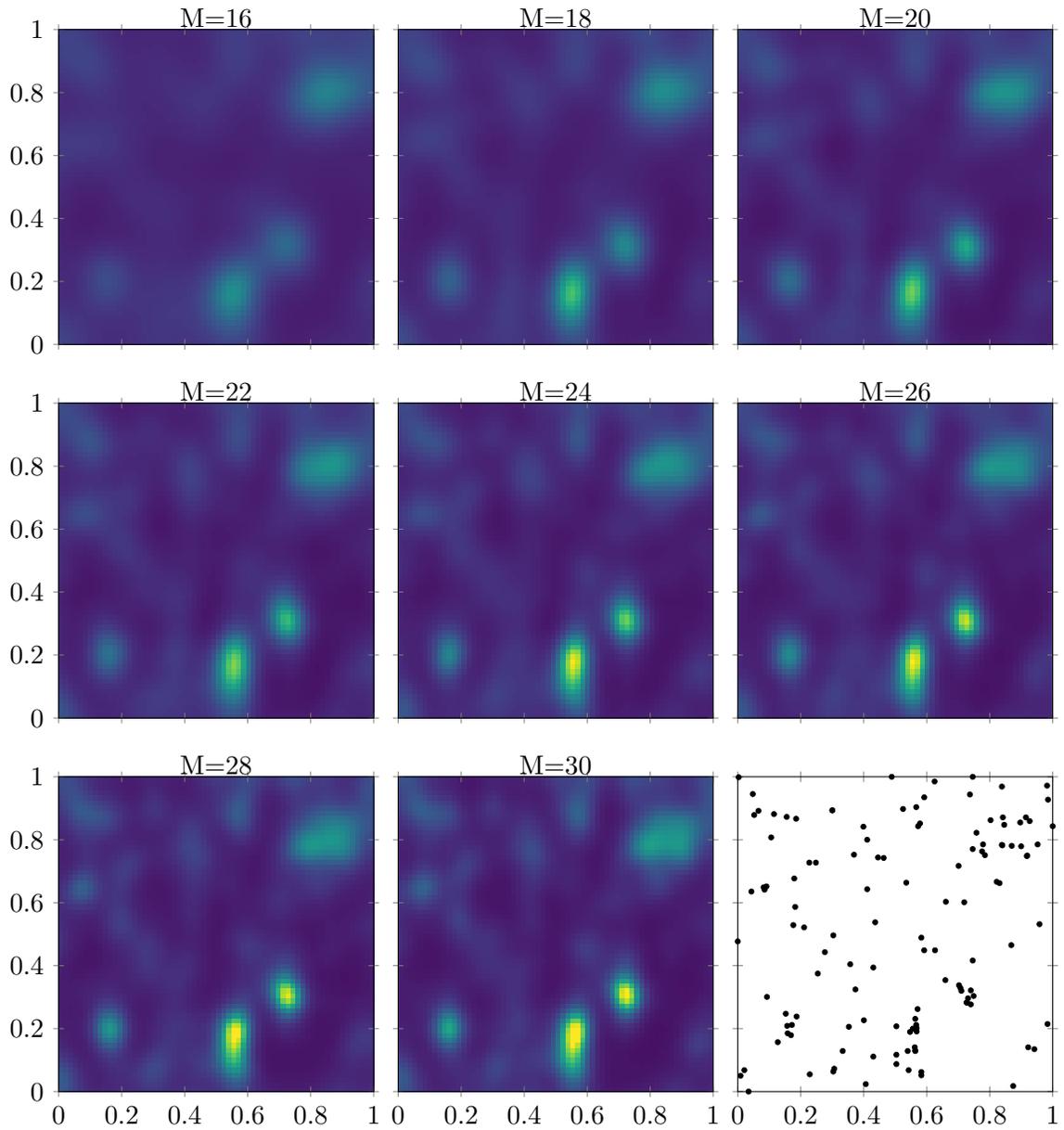

    \setlength\figurewidth{0.4\textwidth}
    \centering
     \caption{\label{fig:pines_intense} The posterior mean of the process intensity with increasing $M$ (denoted above each figure). The spatial locations of the data are shown in the last pane. As $M$ increases, the behaviour of the method stabilizes: we conclude that at $M=30$ there is no benefit from increasing $M$ further. If $M$ is too low, the method is biased towards a longer lengthscale (see also Figure~\ref{fig:pines_lengths}).}
\end{figure}

\begin{figure}
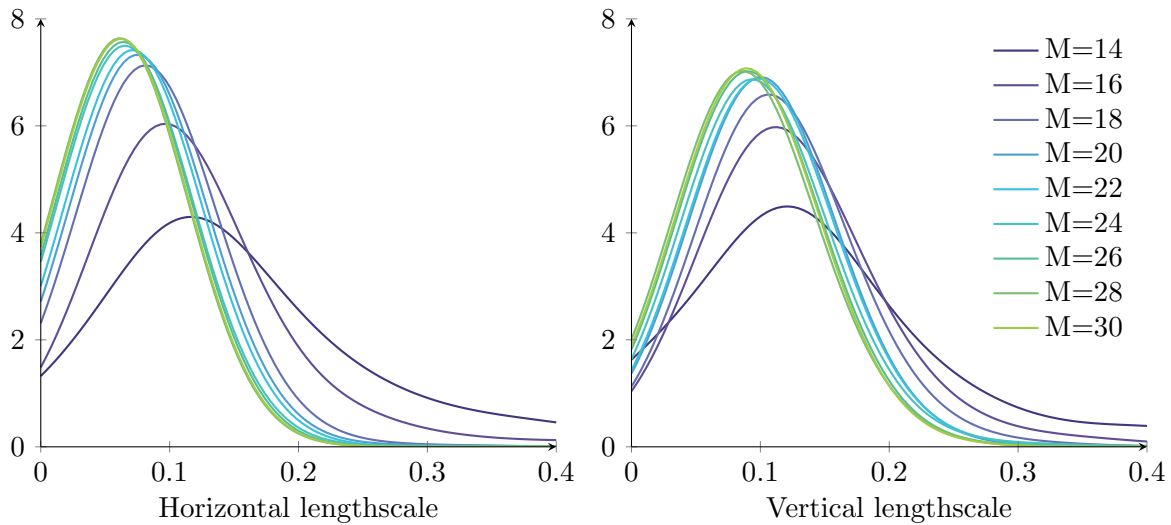

    \setlength\figurewidth{0.4\textwidth}
    \centering%
     \caption{\label{fig:pines_lengths} The posterior distribution of the two lengthscale parameters for the pines problem (see also Figure~\ref{fig:pines_intense}), estimated using kernel density smoothing on the MCMC trace. As more frequencies are used, the posterior distribution converges.}
\end{figure}

To illustrate how the VFF method approaches the posterior in the MCMC case, we ran the HMC algorithm using $M = 14, 16, \ldots, 30$ frequencies. We selected the boundaries of the decomposition $[a, b]$ to be $[-1, 2]^2$, whilst the data were normalized to the unit square $[0, 1]^2$. The posterior mean intensity of the process on the unit square, along with the data, are shown in Figure~\ref{fig:pines_intense}. The figure shows that as $M$ increases, the behaviour of the method stabilizes: there is little to distinguish $M=30$ from $M=28$. If $M$ is too low, then there is a bias towards longer lengthscales, but this is alleviated as $M$ increases. Since more inducing variables must always move the approximation toward the posterior in the KL sense \citep{titsias2009variational}, once the optimal posterior (for each fixed $M$) remains unchanging with an increase in $M$, we must be approximating the posterior closely. A practical approach then, is to use some reasonable number for $M$ (say, 100) and then ensure that the sampled distribution is unmoving with increased $M$. Since our experiments suggest that the lengthscale parameter is particularly sensitive to bias due to $M$ being too low, we suggest examining the convergent behaviour of the lengthscale posterior. Figure~\ref{fig:pines_lengths} shows how the posterior distribution of the lengthscales converges as $M$ increases.

We note that the regular-lattice nature of this problem means that it can also be efficiently solved using fast Fourier transform (FFT) methods \citep[e.g.][]{taylor2014inla}. However, our approach offers several advantages. First, in our method the number of frequencies used is a variational parameter: we have discussed how to select $M$ above. Since the dimension of the vector which is subject to MCMC depends on the number of frequencies, it is desirable to decouple it from the grid density: fine grids are desired to resolve the spatial nature of the process, whilst the length of the vector $\bu$ is desired to be short for computational reasons. Second, our approach does not require the embedding of the kernel matrix into a circulant matrix which can occasionally suffer from non-positive-definiteness. Third and most importantly, our method is not restricted to problems on a regular grid.
 \section{Discussion and future directions}\label{par:discussion}
In this paper we have presented a method which we refer to as the {\em Variational Fourier Features} (VFF) approach to Gaussian process modelling. It combines the variational approach to sparse approximations in Gaussian processes with a representation based on the spectrum of the process. This approach inherits its appealing approximative construction from the variational approach, but with the representative power and computational scalability of spectral representations.

We generalized the method to several dimensions in cases where the kernel exhibits special structure in the form of additive or separable (product) covariance functions. This choice made it possible to apply the scheme to higher-dimensional test problems in the experiments, for which we demonstrate good performance both in terms of computational speed and predictive power. 

Our example application of predicting airline delays suggested that the additive structure worked well for that problem: for data subsets where the full model could be fitted, there was little performance difference between the additive and isotropic case, and indeed the VFF method was competitive with both, suggesting that the posterior was very well approximated.  It seems implausible that this structure could work well for all problems, and clearly it will not work well where interesting interactions occur between covariates. 

The method {\em is} capable of modelling interactions between covariates using separable kernel structures. 
As was noted in the text, the computational scalability in this case limits the method to low-dimensional problems, because the number of inducing frequencies scales exponentially with the input dimension. In this sense this approach is not a sparse approximation, but rather based on a `dense' basis function projection. Like related approximation schemes \citep{solin2014hilbert, wilson2015kernel, nickson2015blitzkriging}, VFF scales well as the number of data increases, with the drawback of difficulty in dealing with high-dimensional problems out-of-the-box. 

To further improve the computational scaling in high-dimensional problems, there are several approaches which could apply. \citet{wilson2015thoughts} presents a projection approach \citep[based on the ideas in][]{wilson2015kernel}. These ideas could be expanded to our approach as well, and we might expect the variational properties of the approximation to be useful in this case. Alternatively, replacing the dense grid spectrum with a set of sparsely well-chosen spectral points is an option that can --- at least for models or data with suitable structure --- provide computational remedy in high input dimensions. Future work might incorporate some ideas from the sparse spectrum method \citep{lazaro2010sparse} where the frequencies are optimized, though we might hope that the variational formulation could prevent over-fitting.

We note that one of the key computations in our method, the multiplication $\Kuf\by$, is precisely a non-uniform FFT \citep[NUFFT, see][]{dutt1993fast, greengard2004accelerating, bernstein2004handbook}, which can be approximated in $\cO(N \log N)$ operations. Although we have so far avoided using this technology in our implementations, some initial experiments suggest that further speed-up is possible at the cost of a very small loss in accuracy. 
In the special case where the data lie on a regular grid, it is possible to compute the product $\Kuf\by$ exactly in $\cO (N \log N)$ operations using an FFT. Again we have not exploited this in our presentation so far, and future work may consider the relations between VFF and FFT methods based on the circulant embedding trick \citep{taylor2014inla, turner2010statistical}. 

A further limitation of our current presentation is that we have only considered \matern covariance functions with half-integer orders. A productive future direction will be to expand the number of kernels which can be decomposed using Variational Fourier Features. An interesting class of kernels may be the spectral mixture kernel \citep{wilson2013gaussian}.  Although that work was based on frequency-shifted versions of the RBF covariance, we anticipate that frequency-shifted \matern kernels would work just as well, and may be more amenable to the VFF framework. 

\add{In our derivations, we have considered uniform input densities on some bounded interval $[a,b]$. Instead of considering such compact bounded subsets of $\mathbb{R}$, it might be possible to extend our methodology to more general input densities \citep[cf.][]{Williams+Seeger:2000}. In special cases this can lead to convenient approximations for the eigen-decomposition of the kernel. For example, for a RBF covariance function and a Gaussian input density, the eigenfunctions can be given in closed-form in terms of Hermite polynomials \citep[see][Ch.~4.3]{williams2006gaussian}. There is a connection between harmonic Fourier approximations like ours and the covariance operator defined through the covariance function and the input density. The connection goes back to Sturm-Liouville theory and is further discussed in \citet{solin2014hilbert}.}

Finally, we note that VFF may be particularly well-suited to machine learning methods where Gaussian processes are embedded in some further model structure, such as the Gaussian process latent variable model \citep[GPLVM,][]{lawrence2005probabilistic} and its variational-Bayesian variant \citep{titsias2010bayesian}, as well as deep Gaussian processes \citep{damianou2013deep}. As described by \citet{damianou2016variational}, variational inference in these models involves propagating uncertainty by convolving the $\Kuf$ matrix with Gaussian approximations to the input distribution. Since the VFF method has precisely sinusoidal $\Kuf$ matrices, this convolution will take a simple form, and we envisage that combining the VFF framework with variational uncertainty propagation will lead to substantial improvements in inference for these models.

\section*{Acknowledgements}\label{par:acknowledgements}
Part of this research was conducted during the invitation of J.~Hensman at Mines Saint-\'Etienne. This visit was funded by the Chair in Applied Mathematics OQUAIDO, which gather partners in technological research (BRGM, CEA, IFPEN, IRSN, Safran, Storengy) and academia (Ecole Centrale de Lyon, Mines Saint-\'Etienne, University of Grenoble, University of Nice, University of Toulouse) on the topic of advanced methods for computer experiments. J.~Hensman gratefully acknowledges a fellowship from the Medical Research Council UK. A.~Solin acknowledges the Academy of Finland grant 308640. We acknowledge the computational resources provided by the Aalto Science-IT project. J.~Hensman would like to thank T. Smith for insightful discussions. The authors would like to thank T.~Bui, A.~Vehtari, S.T.~John and the anonymous reviewers who helped to improve this manuscript.
 
\clearpage
\appendix
\section{$L_2$ Projection on Fourier Features for Mat\'ern-$\frac{1}{2}$}
\label{par:l2ff}
In this Appendix we derive expressions for Fourier features using the $L_2$ norm.
\subsection{Covariance between inducing variable and GP}
\begin{align}
    \cov(u_m, f(x)) &= \mathbb E \left[ u_m f(x) \right] = \int_a^b \mathbb E[f(t) f(x)] e^{-i \omega_m (t-a)} \dee t = \int_a^b k(t, x) e^{-i \omega_m (t-a)} \dee t \\
        &= e^{-i\omega_m a} \sigma^2 \left( \int_a^x e^{\lambda(t-x) + i \omega_m t} \dee t + \int_x^b e^{\lambda(x-t) + i \omega_m t}\dee t\right)\\
    &= e^{-i\omega_m a} \sigma^2 \left( \left[\frac{e^{\lambda(t-x) + i \omega_m t}}{i\omega_m + \lambda}\right]_a^x + \left[\frac{e^{\lambda(x-b) + i \omega_m t}}{i\omega_m - \lambda}\right]_x^b \right)\\
    &= e^{-i\omega_m a} \sigma^2 \left( \frac{e^{i\omega_m x} - e^{\lambda(a-x) + i \omega_m a}}{i\omega_m + \lambda} + \frac{e^{\lambda(x-b) + i\omega_m b} - e^{i\omega_m x}}{i\omega_m - \lambda} \right)\\
    &= \sigma^2 \left( \frac{e^{i\omega_m (x-a)} - e^{\lambda(a-x)}}{i\omega_m + \lambda} + \frac{e^{\lambda(x-b)} - e^{i\omega_m (x-a)}}{i\omega_m - \lambda} \right)\\
    &= \frac{2\sigma^2\lambda}{\lambda^2 + \omega_m^2}e^{i\omega_m (x-a)} + \frac{\sigma^2}{\lambda^2 + \omega_m^2}\left(\lambda[-e^{a-x} - e^{x-b}] + i\omega_m[e^{a-x} - e^{x-b}]\right) \\
    &= s_{1/2}(\omega_m) e^{i\omega_m (x-a)} + s_{1/2}(\omega_m) \frac{1}{2 \lambda}\left(\lambda[-e^{a-x} - e^{x-b}] + i\omega_m[e^{a-x} - e^{x-b}]\right)\,.
\end{align}
\subsection{Covariance between inducing variables}
For the $L_2$ Fourier features, the covariance between two features $u_m$ and $u_{m'}$ depends on the basis functions being sines or cosines. We will thus detail the two cases considering either the real or imaginary part of $\bu$. The following integrals, which can be computed using $\cos(\omega x) = \mathcal{R}\mathrm{e}( e^{i \omega x} )$, will be of particular interest:
 \begin{align}
    \int_x^y e^{\lambda s} \cos(\omega s) \dee s &= \frac{1}{\lambda^2 + \omega^2}
    \big[ e^{\lambda x} (- \lambda \cos(\omega x) - \omega \sin(\omega x))
         +e^{\lambda y} (  \lambda \cos(\omega y) + \omega \sin(\omega y)) \big], \\
    \int_x^y e^{\lambda s} \sin(\omega s) \dee s &= \frac{1}{\lambda^2 + \omega^2}
    \big[ e^{\lambda x} ( \omega \cos(\omega x) - \lambda \sin(\omega x))
         +e^{\lambda y} (-\omega \cos(\omega y) + \lambda \sin(\omega y)) \big]\,.
 \end{align}

\noindent \textbf{case 1:} $i,j \leq M$ (cosine block)
\begin{flalign}
    \cov[u_m,u_{m'}]
    & = \mathbb E \left[\int_a^b f(s) \cos(-\omega_m (s-a))\dee s\, \int_a^b f(t) \cos(-\omega_{m'} (t-a) ) \dee t \right]\\
    & = \int_a^b \int_a^b k(s, t) \cos(-\omega_m (s-a)) \cos(-\omega_{m'} (t-a))\dee s \dee t\\
    & = \int_0^{a-b} \int_0^{b-a} k(s, t) \cos(\omega_m s) \cos(\omega_{m'} t)\dee s \dee t\\
    & =  \frac{1}{\lambda^2 + \omega_m^2} \int_0^{b-a}  \Big[ 2 \lambda  \cos(\omega_m t) - \lambda (e^{ -\lambda t} + e^{\lambda (t+a-b)}) \Big] \cos(\omega_{m'} t) \dee t \\
    & =  \frac{2 \lambda}{\lambda^2 + \omega_m^2} \int_a^b  \Big[ \cos(\omega_m t) - e^{ -\lambda t} \Big] \cos(\omega_{m'} t) \dee t\,.
\end{flalign}
\qquad \textbf{case 1a: $m \neq m'$}
\begin{flalign}
    \cov[u_m,u_{m'}] & = \frac{- 2 \lambda}{\lambda^2 + \omega_m^2} \int_0^{b-a}  e^{-\lambda t} \cos(\omega_{m'} t) \dee t \\
        & = \frac{ - 2 \lambda^2 }{(\lambda^2 + \omega_m^2)(\lambda^2 + \omega_{m'}^2) }
    \big[1-  e^{\lambda (a-b)} \big]\,.
\end{flalign}
\qquad \textbf{case 1b: $m = m' \neq 0$}
\begin{flalign}
    \cov[u_m,u_{m'}] & = \frac{ - 2 \lambda^2 }{(\lambda^2 + \omega_m^2)(\lambda^2 + \omega_{m'}^2) } \big[1-  e^{\lambda (a-b)} \big] + (b-a)\frac{ \lambda }{\lambda^2 + \omega_m^2}\,.
\end{flalign}
\qquad \textbf{case 1c: $m = m' = 0$}
\begin{flalign}
    \cov[u_m,u_{m'}] & = \frac{ - 2 \lambda^2 }{(\lambda^2 + \omega_m^2)(\lambda^2 + \omega_{m'}^2) } \big[1-  e^{\lambda (a-b)} \big] + 2(b-a)\frac{ \lambda }{\lambda^2 + \omega_m^2}\,.
\end{flalign}

\noindent \textbf{case 2:} $m,\, m' > M$ (sine block)
\begin{flalign}
    \cov[u_m,u_{m'}] & = \int_0^{b-a} \int_0^{b-a}  k(s,t) \sin(\omega_m s) \sin(\omega_{m'} t) \dee s \ \dee t \\
        & =  \frac{2 }{\lambda^2 + \omega_m^2} \int_0^{b-a}  \Big[ \lambda \sin(\omega_m t) + \omega_m e^{-\lambda t} \Big] \sin(\omega_{m'} t) \dee t\,.
\end{flalign}
\qquad \textbf{case 2a: $m \neq m'$}
\begin{flalign}
    \cov[u_m,u_{m'}] & = \frac{ 2 \omega_m}{\lambda^2 + \omega_m^2} \int_0^{b-a}  e^{-\lambda t} \sin(\omega_{m'} t) \dee t \\
                & = \frac{ 2 \omega_{m} \omega_{m'} }{(\lambda^2 + \omega_m^2)(\lambda^2 + \omega_{m'}^2) }
    \big[1-  e^{\lambda (a-b)} \big]\,.
\end{flalign}
\qquad \textbf{case 2b: $m = m'$}
\begin{flalign}
    \cov[u_m,u_{m'}] & = \frac{  2 \omega_{m'}^2 }{(\lambda^2 + \omega_m^2)(\lambda^2 + \omega_{m'}^2) } \big[1-  e^{\lambda (a-b)} \big] + (b-a)\frac{ \lambda }{\lambda^2 + \omega_m^2}\,.
\end{flalign}
 \section{\matern inner products}
\label{par:matern_inners}
The expressions of inner products for \matern RKHS on $[a,b]$ can be found in \cite{durrande2016detecting}. We adopt here the following notations to obtain compact expressions: for any function $h \in \cH$, $\mathcal{I}:\ h \rightarrow h$ is the identity operator and $\mathcal{D}:\ g \rightarrow g'$ is the differentiation operator. As a consequence, $(\lambda \mathcal{I} + \mathcal{D})^2(h)$ is a shorthand for $\lambda^2 h + 2  \lambda h' + h''$.

\begin{align}
\text{\maternonetwo: } \PSi{g,h}{\mathcal{H}_{1/2}} & = \frac{1}{2 \lambda \sigma^2} \int_a^{b} (\lambda \mathcal{I} + \mathcal{D})(g) ( \lambda \mathcal{I} + \mathcal{D})(h) \dee t + \frac{1}{\sigma^2} g(a)h(a) \, , \\
 \text{\maternthreetwo: } \PSi{g,h}{\mathcal{H}_{3/2}} & = \frac{1}{4 \lambda^3 \sigma^2} \int_a^{b} (\lambda \mathcal{I} + \mathcal{D})^2(g) ( \lambda \mathcal{I} + \mathcal{D})^2(h) \dee t  \nonumber \\
 & \qquad + \frac{1}{\sigma^2} g(a)h(a) + \frac{1}{ \lambda^2 \sigma^2} g'(a)h'(a) \, , \\
\text{\maternfivetwo: }\PSi{g,h}{\mathcal{H}_{5/2}} & = \frac{3 }{16 \lambda^5 \sigma^2} \int_a^{b} (\lambda \mathcal{I} + \mathcal{D})^3(g) ( \lambda \mathcal{I} + \mathcal{D})^3(h)  \dee t \nonumber  \\
 & \qquad + \frac{9}{8\sigma^2} g(a)h(a) +  \frac{9}{8 \lambda^4 \sigma^2} g(a)''h''(a)  \nonumber \\
 & \qquad + \frac{3 }{\lambda^2 \sigma^2} \left(g'(a)h'(a) +\frac18 g''(a)h(a) + \frac18 g(a)h''(a) \right) \, .
\label{eq:matern_inner}
\end{align}

\subsection{Gram matrix between Fourier features for the exponential kernel}
We detail here the computations of $\Kuu[i,j] = \PS{\phi_i,\phi_j}$ for the exponential kernel. We recall that $\phi_0 = 1$ and that $\phi_i = \cos (\omega_i (x-a))$ and $\phi_{i+M} = \sin (\omega_i (x-a))$ for $i \in \{1,\dots,M\}$. Furthermore, the frequencies $\omega_i$ are harmonic on $[a,b]$.

\textbf{case 1 :} $i,j \leq M$ (cosine block)
\begin{equation}
\begin{split}
	\Kuu [i,j]
	&= \frac{1}{2 \sigma^2 \lambda} \int_0^{b-a}
		(\lambda \cos(\omega_i s) - \omega_i \sin(\omega_i s))
		(\lambda \cos(\omega_j s) - \omega_j \sin(\omega_j s)) \dee s
		+ \frac{1}{\sigma^2}\,.
\end{split}
\end{equation}
The integral is zero for all non-diagonal terms ($i \neq j$). As a consequence, the block of $\Kuu$ associated with the cosine basis functions are $\mathrm{diag}(\alpha_{cos}) + \sigma^{-2}$, where
\begin{equation}
\begin{split}
	\alpha_{cos} [i]
	&= \frac{1}{2 \sigma^2 \lambda} \int_0^{b-a}
		\lambda^2 \cos(\omega_i s)^2 + \omega_i^2 \sin(\omega_i s)^2 \dee s
 =
	\begin{cases}
		\displaystyle \frac{\lambda (b-a)}{2 \sigma^2} & \text{ if } i = 0\,, \\[1em]
		\displaystyle \frac{b-a}{4 \sigma^2 \lambda} (\lambda^2 +  \omega_i^2)  & \text{ if } i \neq 0\,,
	\end{cases}
\end{split}
\end{equation}
which leads to $\alpha_{cos} = \tfrac12 (b-a) \left[ 2 s(0)^{-1}, s(\omega_1)^{-1}, \dots, s(\omega_M)^{-1} \right]$.

\textbf{case 2 :} $i,j > M$ (sine block)
\begin{equation}
\begin{split}
	\Kuu [i,j]
	&= \frac{1}{2 \sigma^2 \lambda} \int_0^{b-a}
		(\lambda \sin(\omega_i s) - \omega_i \cos(\omega_i s))
		(\lambda \sin(\omega_j s) - \omega_j \cos(\omega_j s)) \dee s\,.
\end{split}
\end{equation}
The $\Kuu$ block associated to the sine basis functions is exactly a diagonal matrix with:
\begin{equation}
\begin{split}
	\Kuu [i,i]
	&= \frac{1}{2 \sigma^2 \lambda} \int_0^{b-a}
		(\lambda^2 \sin(\omega_i s)^2 + \omega_i^2 \cos(\omega_i s)) \dee s
		= \frac{b-a}{4 \sigma^2 \lambda} (\lambda^2 +  \omega_i^2)\,.
\end{split}
\end{equation}
Similarly to the cosine block, we can write the sine block as $\mathrm{diag}(\alpha_{sin})$ with $\alpha_{sin} = \tfrac12 (b-a) \left[s(\omega_1)^{-1}, \dots, s(\omega_M)^{-1} \right]$.

\textbf{case 3 :} $i \leq 2M +1 < j $ (off-diagonal block) leads to $\Kpp [i,j] = 0$.

\subsection{Gram matrix associated to Fourier features for the Mat\'ern-$\frac{3}{2}$ kernel}

\textbf{case 1 :} $i,j \leq M$ (cosine block)\\

\begin{equation}
\begin{split}
	\Kpp [i,j] &= \frac{1}{4 \sigma^2 \lambda^3} \int_0^{b-a}
		(\lambda^2 \cos(\omega_i s) - 2 \lambda \omega_i \sin(\omega_i s) - \omega_i^2 \cos(\omega_i s)) \times \\
	&	\qquad \qquad \qquad \qquad
		(\lambda^2 \cos(\omega_j s) - 2 \lambda \omega_j \sin(\omega_j s) - \omega_j^2 \cos(\omega_j s))
		\dee s \\
	&	\qquad + \frac{1}{\sigma^2} \cos(\omega_i a) \cos(\omega_j a) +  \frac{1}{\lambda^2 \sigma^2} \omega_i \omega_j \sin(\omega_i a) \sin(\omega_j a) \\
	& =
	\begin{cases}
		\displaystyle \frac{1}{\sigma^2} & \text{ if } i \neq j\,,\\[1em]
		\displaystyle \frac{1}{4 \sigma^2 \lambda^3} \int_0^{b-a} \lambda^4 \cos^2(\omega_i s) +  \omega_i^4 \cos^2(\omega_i s) \\
		\displaystyle \qquad \qquad - 2\lambda^2 \omega^2 \cos^2(\omega_i s) + 4 \lambda^2 \omega_i^2 \sin^2(\omega_i s) ds + \frac{1}{\sigma^2} & \text{ if } i = j\,,
	\end{cases}\\
	& =
	\begin{cases}
		\displaystyle \frac{1}{\sigma^2} & \text{ if } i \neq j\,,\\[1em]
		\displaystyle \frac{b-a}{8 \sigma^2 \lambda^3} (\lambda^2 +  \omega_i^2)^2 + \frac{1}{\sigma^2} & \text{ if } i = j \neq 0\,,\\[1em]
		\displaystyle \frac{\lambda (b-a)}{4 \sigma^2} + \frac{1}{\sigma^2} & \text{ if } i = j = 0\,.
	\end{cases}
\end{split}
\end{equation}

\textbf{case 2 :} $i,j > M$ (sine block)
\begin{equation}
\begin{split}
	\Kpp [i,j] &= \frac{1}{4 \sigma^2 \lambda^3} \int_0^{b-a}
		(\lambda^2 \sin(\omega_i s) + 2 \lambda \omega_i \cos(\omega_i s) - \omega_i^2 \sin(\omega_i s)) \times \\
	&	\qquad \qquad \qquad \qquad
		(\lambda^2 \sin(\omega_j s) + 2 \lambda \omega_j \cos(\omega_j s) - \omega_j^2 \sin(\omega_j s))
		\dee s \\
	&	\qquad + \frac{1}{\sigma^2} \sin(\omega_i a) \sin(\omega_j a) +  \frac{1}{\lambda^2 \sigma^2} \omega_i \omega_j \cos(\omega_i a) \cos(\omega_j a) \\
	& =
	\begin{cases}
		\displaystyle \frac{\omega_i\omega_j}{\lambda^2 \sigma^2} & \text{ if } i \neq j\,,\\[1em]
		\displaystyle \frac{1}{4 \sigma^2 \lambda^3} \int_0^{b-a} \lambda^4 \sin^2(\omega_i s) +  \omega_i^4 \sin^2(\omega_i s) \\
		\displaystyle \qquad \qquad - 2\lambda^2 \omega^2 \sin^2(\omega_i s) + 4 \lambda^2 \omega_i^2 \cos^2(\omega_i s) ds + \frac{\omega_i\omega_j}{\lambda^2 \sigma^2} & \text{ if } i = j\,,
	\end{cases}\\
	& =
	\begin{cases}
		\displaystyle \frac{\omega_i\omega_j}{\lambda^2 \sigma^2} & \text{ if } i \neq j\,,\\[1em]
		\displaystyle \frac{b-a}{8 \sigma^2 \lambda^3} (\lambda^2 +  \omega_i^2)^2 + \frac{\omega_i\omega_j}{\lambda^2 \sigma^2} & \text{ if } i = j \,.
	\end{cases}
\end{split}
\end{equation}

\textbf{case 3:} $i \leq M < j $ (off-diagonal block)\\
\begin{equation}
\begin{split}
	\Kpp [i,j] &= \frac{1}{4 \sigma^2 \lambda^3} \int_0^{b-a}
		(\lambda^2 \cos(\omega_i s) - 2 \lambda \omega_i \sin(\omega_i s) - \omega_i^2 \cos(\omega_i s)) \times \\
	&	\qquad \qquad \qquad \qquad
		(\lambda^2 \sin(\omega_j s) + 2 \lambda \omega_j \cos(\omega_j s) - \omega_j^2 \sin(\omega_j s))
		\dee s \\
	&	\qquad + \frac{1}{\sigma^2} \cos(\omega_i a) \sin(\omega_j a) - \frac{1}{\lambda^2 \sigma^2} \omega_i \omega_j \sin(\omega_i a) \cos(\omega_j a) \\
	& =
	\begin{cases}
		\displaystyle 0 & \text{ if } i \neq j\,,\\[1em]
		\displaystyle \frac{1}{4 \sigma^2 \lambda^3} \int_0^{b-a} 2 \lambda^3 \omega_i \cos^2(\omega_i s)  -2 \lambda \omega_i^3 \cos^2(\omega_i s) \\
		\displaystyle \qquad \qquad - 2 \lambda^3 \omega_i \sin^2(\omega_i s)   + 2 \lambda \omega_i^3 \sin^2(\omega_i s) \dee s & \text{ if } i = j\,,
	\end{cases}\\
	& = 0
\end{split}
\end{equation}

\subsection{Gram matrix of Fourier features for the Mat\'ern-$\frac{5}{2}$ kernel}
\begin{equation}
\PSi{g,h}{\mathcal{H}_{5/2}} = \frac{3}{16 \lambda^5 \sigma^2} \int_0^{b-a} L_t (g)L_t (h) \dee t + G(a)\,,
\end{equation}

\begin{equation}
\begin{split}
\text{where  } L_t(g) & = \left( \lambda^3 g(t) + 3 \lambda^2 g'(t) + 3 \lambda g''(t) + g'''(t) \right)\,, \\
 G(a) &= \frac{9}{8\sigma^2} g(a)h(a) +  \frac{9}{8 \lambda^4 \sigma^2} g(a)''h''(a)  \\
 & \qquad  + \frac{3 }{\lambda^2 \sigma^2} \left(g'(a)h'(a) +\frac18 g''(a)h(a) + \frac18 g(a)h''(a) \right)\,.
 \label{eq:IP52}
\end{split}
\end{equation}

\textbf{case 1:} $i,j \leq M$ (cosine block)\\
\begin{equation}
\begin{split}
L_x(\cos(\omega_i .)) &=  \lambda^3 \cos( \omega_i x) - 3 \omega_i \lambda^2 \sin( \omega_i x) - 3 \lambda \omega_i^2 \cos(\omega_i x) + \omega_i^3 \sin(\omega_i x)\\
&=  (\lambda^3 - 3 \lambda \omega_i^2 ) \cos( \omega_i x) + (\omega_i^3 - 3 \omega_i \lambda^2) \sin( \omega_i x)\\
 G(a) &= \frac{9}{8\sigma^2} +  \frac{9}{8 \lambda^4 \sigma^2} \omega_i^2 \omega_j^2 - \frac{3 }{8 \lambda^2 \sigma^2} (\omega_i^2 + \omega_j^2)\\
 &= \frac{1}{\sigma^2} +  \frac{1}{8 \sigma^2}\left(\frac{3\omega_i^2}{\lambda^2} - 1\right)\left(\frac{3\omega_j^2}{\lambda^2} - 1\right)\,.
\end{split}
\end{equation}

\begin{equation}
\begin{split}
	\Kpp [i,j] &=
	\begin{cases}
		\displaystyle G(a) & \text{ if } i \neq j\,,\\[1em]
		\displaystyle \frac{3(b-a)}{32 \lambda^5 \sigma^2} (\lambda^6 - 6 \lambda^4 \omega_i^2 + 9 \lambda^2 \omega_i^4 + \omega_i^6 - 6 \lambda^2 \omega_i^4 + 9 \omega_i^2 \lambda^4) + G(a) & \text{ if } i = j \neq 0\,,\\[1em]
		\displaystyle \frac{3 (b-a)}{16 \lambda^5 \sigma^2}  (\lambda^6 - 6 \lambda^4 \omega_i^2 + 9 \lambda^2 \omega_i^4) + G(a) & \text{ if } i = j = 0\,,
	\end{cases}
\end{split}
\end{equation}
which boils down to
\begin{equation}
\begin{split}
	\Kpp [i,j] &=
	\begin{cases}
		\displaystyle G(a) & \text{ if } i \neq j\,,\\[1em]
		\displaystyle \frac{3(b-a)}{32 \lambda^5 \sigma^2} (\lambda^2 + \omega_i^2)^3 + G(a) & \text{ if } i = j \neq 0\,,\\[1em]
		\displaystyle  \frac{3 \lambda (b-a)}{16 \sigma^2} + G(a) & \text{ if } i = j = 0\,.
	\end{cases}
\end{split}
\end{equation}

\textbf{case 2:} $i,j > M$ (sine block)\\
\begin{equation}
\begin{split}
L_x(\sin(\omega_i .)) &=  \lambda^3 \sin( \omega_i x) + 3 \omega_i \lambda^2 \cos( \omega_i x) - 3 \lambda \omega_i^2 \sin(\omega_i x) - \omega_i^3 \cos(\omega_i x)\\
&=  (\lambda^3 - 3 \lambda \omega_i^2 ) \sin( \omega_i x) - (\omega_i^3 - 3 \omega_i \lambda^2) \cos( \omega_i x)\,,\\
 G(a) &= \frac{3 \omega_i \omega_j}{\lambda^2 \sigma^2}\,,
\end{split}
\end{equation}

\begin{equation}
\begin{split}
	\Kpp [i,j] &=
	\begin{cases}
		\displaystyle G(a) & \text{ if } i \neq j\,,\\[1em]
		\displaystyle \frac{3(b-a)}{32 \lambda^5 \sigma^2} (\lambda^2 + \omega_i^2)^3 + G(a) & \text{ if } i = j \neq 0\,.
	\end{cases}
\end{split}
\end{equation}

\textbf{case 3:} $i \leq M < j $ (off-diagonal block). As previously, \remove{calculation gives} \add{calculations give} $\Kpp [i,j] = 0$. \\
 
\clearpage
\phantomsection
\addcontentsline{toc}{section}{References}
\bibliography{references}

\end{document}